\documentclass[lettersize,journal]{IEEEtran}
\usepackage{amsmath,amsfonts}
\usepackage{algorithmic}
\usepackage{algorithm}
\usepackage{array}
\usepackage[caption=false,font=normalsize]{subfig}
\usepackage{textcomp}
\usepackage{stfloats}
\usepackage{url}
\usepackage{verbatim}
\usepackage{graphicx}
\usepackage{cite}

\usepackage{booktabs}
\usepackage{multirow}
\usepackage{amssymb}

\begin{document}

\title{Fine-Grained Model Merging via Modular Expert Recombination}

% \author{Anonymous Authors}
\author{Haiyun~Qiu, Xingyu~Wu*, Liang~Feng,~\IEEEmembership{Senior Member,~IEEE}, Kay~Chen~Tan,~\IEEEmembership{Fellow,~IEEE}
\thanks{* The corresponding author.}
\thanks{Haiyun Qiu, Xingyu Wu, and Kay Chen Tan are with the Department of Data Science and Artificial Intelligence, The Hong Kong Polytechnic University, Hong Kong, SAR 999077, China. (E-mails: haiyun.qiu@connect.polyu.hk, xingy.wu@polyu.edu.hk, kaychen.tan@polyu.edu.hk)}% <-this % stops a space
\thanks{Liang Feng is with the College of Computer Science, Chongqing University, Chongqing 400044, China. (E-mail: liangf@cqu.edu.cn)}}

        % <-this % stops a space

\markboth{Journal of \LaTeX\ Class Files,~Vol.~14, No.~8, August~2021}%
{Shell \MakeLowercase{\textit{et al.}}: A Sample Article Using IEEEtran.cls for IEEE Journals}

\IEEEpubid{0000--0000/00\$00.00~\copyright~2021 IEEE}

\maketitle

\begin{abstract}
Model merging constructs versatile models by integrating task-specific models without requiring labeled data or expensive joint retraining. Although recent methods improve adaptability to heterogeneous tasks by generating customized merged models for each instance, they face two critical limitations. First, the instance-specific merged models lack reusability, restricting the exploitation of high-quality merging configurations and efficient batch inference. Second, these methods treat each task-specific model as a monolithic whole, overlooking the diverse mergeability of homologous components such as attention and multilayer perceptron layers, and the differing merging sensitivities across components. To address these limitations, we propose MERGE (\underline{M}odular \underline{E}xpert \underline{R}ecombination for fine-\underline{G}rained m\underline{E}rging), a method that enables component-wise model merging and input-aware, on-demand module recombination at inference. MERGE formulates component-wise merging as a bi-objective optimization problem that balances cross-task performance and storage efficiency, and develops a surrogate-assisted evolutionary algorithm to efficiently identify Pareto-optimal merging configurations. These high-quality configurations underpin a reusable modular expert library, from which a lightweight routing network dynamically activates and recombines modular experts to assemble input-specific models and enable efficient inference under storage constraints. Extensive experiments across various model scales, task types, and fine-tuning strategies demonstrate that MERGE consistently outperforms strong baselines and generalizes effectively. %Further interpretability and ablation analyses provide insights into the mechanisms underlying its performance gains.

\end{abstract}


\section{Introduction}

\IEEEPARstart{T}{he} pretrain-finetune paradigm \cite{dodge2020fine,zhang2025LoRA} has become the standard methodology for adapting deep learning models to a broad spectrum of downstream tasks, leading to a rapid proliferation of both pre-trained and finetuned models across diverse domains such as computer vision and natural language processing \cite{wolf2019huggingface}. While these task-specific models achieve remarkable performance on their respective target tasks, deploying multiple models in multi-task scenarios incurs substantial storage overhead and increases deployment complexity, especially for large-scale models \cite{huang2024emr}. Recently, model merging \cite{yang2024model, yadavsurvey, Li2025fusion} has attracted considerable attention as an efficient approach for constructing models with multiple capabilities by integrating existing task-specific models to tackle diverse tasks. This approach offers a training-free alternative to traditional multi-task learning (MTL) \cite{zhang2021survey}, effectively mitigating the high training cost and data privacy concerns associated with joint training. 

\begin{figure}[t]
\captionsetup[sub]{labelfont={normalfont,small}}
\centering
\subfloat[]{%
    \includegraphics[width=\linewidth]{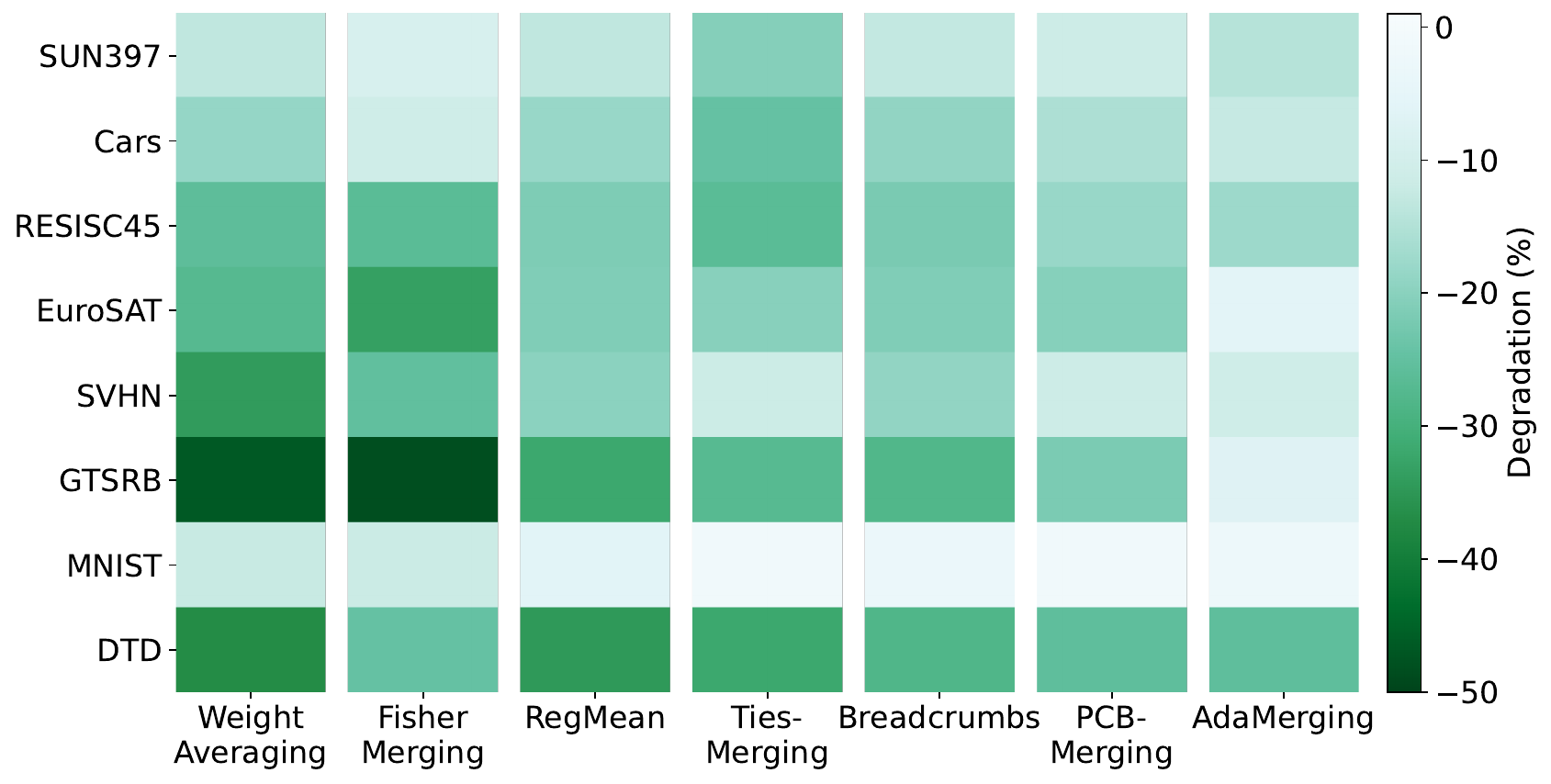}%
    \label{fig1:sub1}}
\vfill
\subfloat[]{%
    \includegraphics[width=0.9\linewidth]{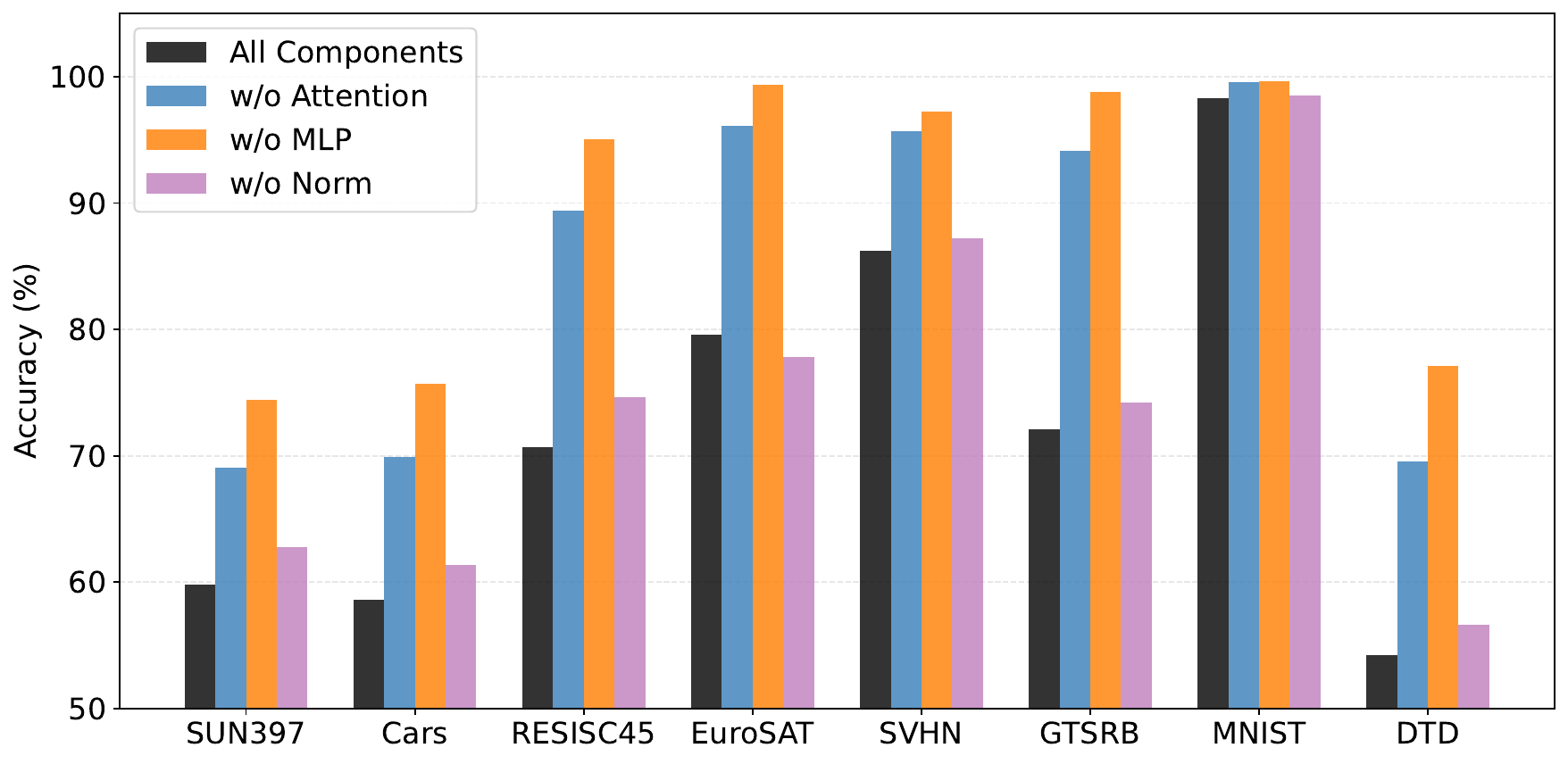}%
    \label{fig1:sub2}}
\caption{Exploratory analyses underscore the necessity of fine-grained merging. (a) Existing methods that generate input-agnostic models exhibit substantial performance degradation compared to task-specific models, highlighting the importance of input-aware merging. Darker colors indicate greater degradation. (b) Component-wise comparisons reveal heterogeneous merging sensitivities across components, emphasizing the need for component-wise merging.}
\label{fig:motivate}
\end{figure}

Traditional model merging methods primarily aim to construct a unified model capable of handling all tasks. The most straightforward technique involves weighted averaging of model parameters fine-tuned on different tasks \cite{wortsman2022model}. While easy to implement, such simple averaging often leads to severe performance degradation \cite{du2024parameter,zhou2024hm3}. To improve the performance of merged models, more advanced merging strategies have been proposed \cite{jin2022dataless, ilharco2022editing, yadav2023ties, yang2023adamerging,gu2023hierarchical}. Nevertheless, these methods still fundamentally generate input-agnostic merged models by combining all task-specific models within a fixed and limited parameter space. Such indiscriminate, coarse-grained merging inevitably dilutes expert knowledge and limits adaptability to heterogeneous tasks, resulting in a substantial performance gap between the merged model and task-specific models, as illustrated in Fig.~\ref{fig:motivate}\subref{fig1:sub1}.

\IEEEpubidadjcol
To enhance the adaptability of model merging, recent studies have explored input-aware merging strategies that dynamically generate customized merged models for each input \cite{huang2024emr, lu2024twin, ye2025dynamic}. Inspired by the Mixture-of-Experts (MoE) architecture \cite{cai2025survey}, these methods store task-specific model parameters as experts, which are dynamically activated during inference to construct a unique merged model on a per-instance basis. However, this line of work exposes a fundamental tension between adaptability and reusability. Existing methods predominantly prioritize adaptability at the cost of low reusability and substantial inefficiency. Specifically, while instance-specific merging enables fine-grained adaptation at instance level, it produces disposable merged models that cannot be reused, cached, or shared across inputs, thereby limiting the exploitation of high-quality merging configurations. Moreover, instance-specific inference fail to leverage batch processing efficiency.

Another key limitation is the exclusive focus on input-aware merging, with little attention paid to component-wise granularity. Current approaches treat each task-specific model as a monolithic whole, failing to capture the diverse mergeability of homologous model components. For instance, when task-specific models are semantically aligned in their attention layers, these components can often be merged with minimal information loss. In contrast, multilayer perceptron (MLP) layers tend to encode highly task-specific transformations, making them far more prone to destructive interference during merging. This component-wise heterogeneity is not anecdotal but systematic. As shown in Fig.~\ref{fig:motivate}\subref{fig1:sub2}, vision models exhibit a clear sensitivity hierarchy: MLP layers are the most sensitive to merging, followed by attention layers, while normalization layers have negligible impact.

To address the aforementioned limitations, we propose MERGE (\underline{M}odular \underline{E}xpert \underline{R}ecombination for fine-\underline{G}rained m\underline{E}rging), a novel approach that enables both component-wise and input-aware fine-grained model merging through modular expert recombination. Specifically, MERGE first decomposes task-specific models into functional components, laying the groundwork for component-wise merging. To reconcile input adaptability with reusability, a natural solution is to pre-construct and store a set of high-quality merging configurations that can be selectively invoked during inference based on the input. Since retaining more expert information may improve fidelity and performance but also substantially increase storage overhead \cite{ye2025dynamic}, the component-wise merging process in MERGE is formulated as a bi-objective optimization problem, jointly balancing cross-task performance and storage efficiency. 

Achieving component-wise merging through bi-objective optimization is highly challenging due to the combinatorial explosion in a high-dimensional, discrete search space, where the number of possible configurations increases exponentially with the number of task-specific models and components. Additionally, each configuration requires expensive evaluation across multiple tasks. This combinatorial and discrete nature, coupled with expensive evaluations, renders both brute-force and gradient-based optimization approaches fundamentally infeasible. To efficiently identify Pareto-optimal component-wise merging configurations, we develop a surrogate-assisted bi-objective evolutionary algorithm. The surrogate model acts as a lightweight performance predictor, obviating a significant portion of expensive evaluations and thereby accelerating the search for high-quality configurations. To further enhance storage efficiency in fine-grained merging, MERGE incorporates a merge-aware conditional quantization strategy, which leverages post-merging parameter distribution shifts to identify compressible units, enabling simultaneous compression of parameter size and precision during optimization. Ultimately, the Pareto-optimal configurations obtained from optimization underpin the construction of a modular expert library composed of diverse, reusable modules. At inference time, guided by a lightweight routing network, MERGE dynamically assembles input-specific models by activating and recombining appropriate modular experts from the constructed library, adhering to user-specified storage constraints and flexibly adapting to the diverse requirements of heterogeneous tasks.

In summary, MERGE front-loads component-wise model merging to an offline optimization stage and enables input-aware, on-demand reuse of modular experts at inference time. By systematically orchestrating fine-grained component recombination with a library of reusable, high-quality expert configurations, MERGE provides a principled solution to addressing the longstanding complexity in model merging. Our contributions are summarized as follows:
%of achieving component-wise flexibility, input-aware adaptability, cross-task performance, and storage efficiency 
\begin{enumerate}
    \item \textbf{Input-Aware Adaptability:} We construct a modular expert library with input-aware reuse, which overcomes the limitations of non-reusability and inference inefficiency in existing instance-specific dynamic merging methods by enabling the reuse of high-quality merging configurations across different inputs.
    \item \textbf{Component-Wise Flexibility:} We propose an adaptive component-wise merging paradigm that explicitly leverages the functional heterogeneity of model components, allowing MERGE to identify high-quality merging configurations while avoiding suboptimal combinations.
    \item \textbf{Cross-Task Ability with Storage Efficiency:} We formulate component-wise merging as a bi-objective optimization problem and develop a surrogate-assisted evolutionary algorithm to efficiently discover a diverse set of Pareto-optimal merging configurations, offering a controllable trade-off between cross-task performance and storage efficiency for resource-aware deployment.
    %\item Through systematic analysis, we uncover previously unexplored phenomena in fine-grained model merging, such as the diminishing marginal utility of component-wise merging under increasingly relaxed storage constraints, and distinct component-wise integration preferences across model architectures, providing new insights into the intrinsic mechanisms of model merging.
\end{enumerate}

The rest of the paper is organized as follows. Section~\ref{sec:Preliminary and Related Work} reviews the most relevant research on model merging. Section~\ref{sec:Proposed Method} elaborates on the core procedures of the proposed MERGE method. In Section~\ref{sec:Experiments}, we report comprehensive experimental results and provide in-depth analyses. Finally, conclusions are drawn in Section~\ref{sec:Conclusion}.

\section{Related Work}
\label{sec:Preliminary and Related Work}

In the pretrain-finetune paradigm, pre-trained models are fine-tuned for specific downstream tasks. Supervised fine-tuning strategies can be broadly classified into: (1) Full Fine-Tuning (FFT), which updates all parameters of the pre-trained model, and (2) Parameter-Efficient Fine-Tuning (PEFT) \cite{han2024parameter,zhang2025LoRA}, which modifies only a small set of inserted modules while keeping the majority of the model weights frozen.

Consider a pre-trained model parameterized by weights $\theta_{\text{pre}}$, which is fine-tuned on a downstream task $t \in \{1,2, \dots, T\}$ to obtain a task-specific model $\theta_t$. Model merging aims to construct a merged model $\theta_m$ with integrated capabilities by combining existing task-specific models $\{\theta_t\}_{t=1}^T$ in a training-free manner. This paradigm has demonstrated significant potential and broad applicability \cite{yang2024model,yadavsurvey, wu2024evolutionary}. Depending on whether merging is performed prior to or during inference, existing techniques can be broadly categorized into static merging and dynamic merging. 

Static merging refers to combining multiple task-specific models into a single unified model intended to handle diverse tasks. Early works primarily merged models fine-tuned on the same task to improve the performance of the final models \cite{rame2023model,rame2022diverse}. With the development of the pretrain–finetune paradigm, recent efforts have shifted toward merging models specialized for different tasks. A straightforward approach is weighted averaging. For instance, simple weight averaging \cite{wortsman2022model} uniformly combines all source model parameters as:
\begin{equation}
    \theta_m = \frac{1}{T}\sum_{t=1}^{T}\theta_t.
\end{equation}
Despite its simplicity, this method often leads to notable performance drops. Consequently, more sophisticated approaches have been developed, such as Fisher-Merging \cite{matena2022merging} and RegMean \cite{jin2022dataless}, which compute weighting coefficients using Fisher matrices and inner-product matrices, respectively. 

Beyond directly averaging model weights, Task Arithmetic \cite{ilharco2022editing} introduces task vectors $\tau_t = \theta_t - \theta_{\text{pre}}$ to quantify the difference between task-specific and pre-trained models, which enables merging via scaled vector addition: 
\begin{equation}
    \theta_m = \theta_{\text{pre}} + \lambda\sum_{t=1}^T\tau_t
\end{equation}
where ${\lambda}$ is a predefined global merging coefficient. Building on this concept, subsequent methods employ post-hoc pruning strategies to mitigate task interference by exploiting the inherent redundancy of task vectors. For instance, Ties-Merging \cite{yadav2023ties} mitigates interference by trimming low-magnitude parameters and resolving sign conflicts, while Breadcrumbs \cite{davari2024model} prunes both small values and parameter outliers. DARE \cite{yu2024language} randomly drops most elements and rescales the remaining ones, and PCB-Merging \cite{du2024parameter} evaluates parameter importance via a competitive balance strategy, thereby dropping unimportant parameters and rescaling crucial ones. In contrast to most static merging methods that rely on manually tuning a single global merging coefficient, AdaMerging \cite{yang2023adamerging} learns per-task or per-layer merging coefficients through unsupervised training. While static merging methods are storage-efficient, their one-size-fits-all merged models are input-agnostic, which limits their adaptability to diverse task requirements. As a result, they often suffer from significant performance degradation relative to individual task-specific models or MTL models, especially as the number of tasks grows \cite{yang2024model}. 

To bridge the performance gaps, dynamic merging generates a customized merged model for each input by integrating shared and task-specific parameters during inference:
\begin{equation}
    \theta_m(x) = \theta_{s} + \sum_{t=1}^T\lambda_t(x)M_t
\end{equation}
where $\theta_{s}$ denotes the shared parameters across all task-specific models, $M_t$ is the additionally stored task-specific parameters, and $\lambda_t(x)$ is a weight predicted by an auxiliary routing network. For example, EMR-Merging \cite{huang2024emr} constructs a unified vector and derives a task-specific mask and rescaler for each task vector. However, EMR-Merging assumes perfect routing, where each expert exclusively handles inputs from its designated task, limiting its ability to process data with unknown task labels. In contrast, Twin-Merging \cite{lu2024twin} designs a dynamic routing mechanism and merges the shared model with compressed expert parameters during inference. Despite their notable performance improvements over static methods, existing dynamic merging approaches still face two major challenges. First, instance-specific merging and inference produce non-reusable merged models, resulting in low reusability of high-quality merging configurations and inefficient inference. Second, they typically overlook component-wise heterogeneity \cite{meng2022locating, geva2021transformer} by treating each task-specific model as a monolithic whole, thereby failing to capture complex interactions among homologous model components and the heterogeneous sensitivities of different components to merging.

Overall, despite significant progress in model merging, existing research has yet to jointly account for the inherent heterogeneity of both model components and inputs, and still faces limitations such as low model reusability and inefficient inference. 

\begin{figure*}[t]
    \centering
    \includegraphics[width=\textwidth, height = 9.5cm]{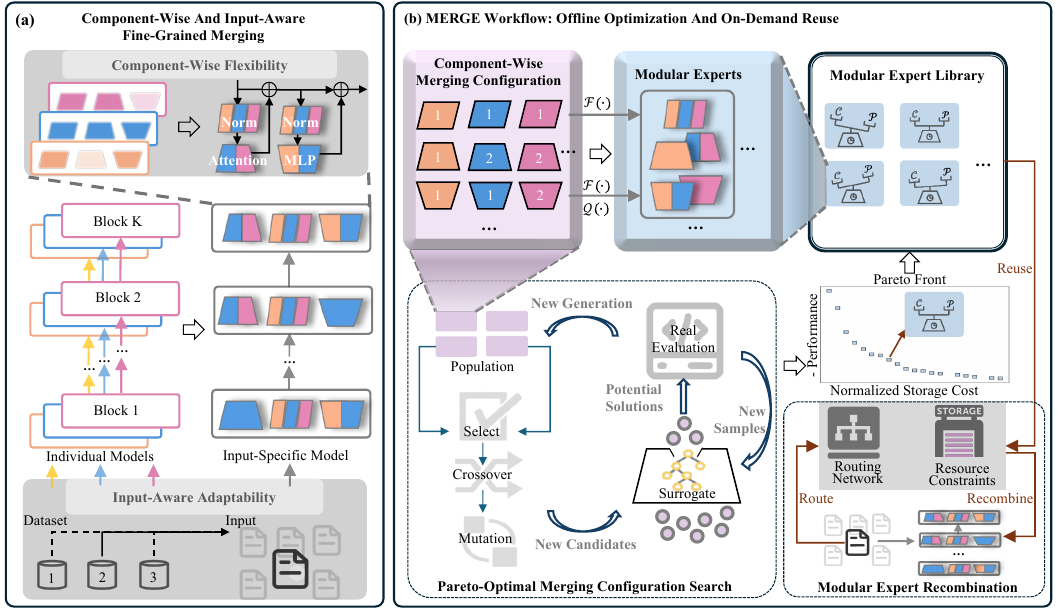}
    \caption{Overview of MERGE. (a) MERGE jointly achieves component-wise flexibility and input-aware adaptability in fine-grained merging. Component-wise merging obtains high-quality, reusable modular experts through offline merging of homologous components, while input-aware module recombination adapts merged models to various inputs during inference. (b) MERGE workflow. During offline optimization, MERGE introduces surrogate-assisted bi-objective evolutionary optimization to identify Pareto-optimal component-wise merging configurations that balance performance and storage cost. These configurations underpin the construction of a reusable modular expert library. During on-demand reuse at inference, guided by a lightweight routing network, MERGE recombines modular experts from this library to assemble input-specified models under user-specified resource constraints.}
    \label{fig:main-framework}
\end{figure*}

\section{Proposed Method}
\label{sec:Proposed Method}

In this study, we propose a method that achieves fine-grained merging along both component-wise and input-aware dimensions. As illustrated in Fig.~\ref{fig:main-framework}(a), the component-wise dimension adaptively merges homologous components from multiple task-specific models into reusable modular experts, while the input-aware dimension dynamically assembles specialized merged models by recombining appropriate modular experts during inference. In the following, we first outline the overall framework of the proposed MERGR in Section~\ref{subsec:Overview of MERGE}, which centers on two key stages, offline optimization, implementing component-wise merging, and on-demand reuse, enabling input-aware module recombination. Subsequently, to clarify the offline optimization stage, we define the component-wise merging configuration representation in Section~\ref{subsec:Fine-grained Solution Representation}, and introduce the Pareto-optimal merging configuration search process in Section~\ref{subsec:Pareto-Optimal Solution Search}. Finally, we detail the on-demand reuse stage for modular expert recombination in Section~\ref{subsec:On-Demand Modular Expert Recombination}.

\subsection{Overview of MERGE}
\label{subsec:Overview of MERGE}

MERGE comprises two interrelated stages, offline optimization and on-demand reuse, as illustrated in Fig.~\ref{fig:main-framework}(b). Offline optimization is performed once to effectively derive reusable modular experts, while on-demand reuse leverages these modular experts to build deployment-time merged models tailored to specific inputs under user-specified storage constraints.

The offline optimization stage is designed to achieve component-wise flexibility by identifying high-quality merging configurations for homologous components across multiple task-specific models, and subsequently consolidating the resulting modules into a reusable modular expert library. As outlined in Algorithm~\ref{alg: offline_optimization}, the procedure begins by decomposing the pre-trained and task-specific models into functional components (line 1). To facilitate precise control at the component level, we define the component-wise merging configuration representation. Building upon this, the merging configuration search process is formulated as a bi-objective optimization problem to balance the trade-off between cross-task performance and storage efficiency. To efficiently explore the vast, discrete search space and reduce the reliance on expensive evaluations, we develop a surrogate-assisted evolutionary algorithm tailored for searching Pareto-optimal merging configurations (lines 2-16). Ultimately, these identified high-quality configurations are utilized to construct the reusable modular expert library (line 17). 

\begin{algorithm}[tb]
\caption{Offline Optimization in MERGE}
\label{alg: offline_optimization}
\textbf{Input}: Pre-trained model $\theta_{\text{pre}}$, task-specific models $\{\theta_t\}_{t=1}^T$, validation sets $\{ \mathcal{D}_t^\prime \}_{t=1}^{T}$\\
\textbf{Output}: Pareto-optimal set $\mathcal{G}^\ast$, modular expert library $\mathbb{L}$
\begin{algorithmic}[1] %[1] enables line numbers
\STATE $\{\theta_{\text{pre}}^l\}_{l=1}^L, \{\theta_t^l\}_{t=1, l=1}^{T,L} \leftarrow$ Decompose $\theta_{\text{pre}}$ and $\{\theta_t\}_{t=1}^T$ into functional components;
\STATE Initialize the population $P_0$, Pareto-optimal set $\mathcal{G}^\ast \leftarrow \{ \mathbf{G} \in P_0 \mid \nexists \mathbf{G}^\prime \in P_0 \text{ such that } \mathbf{G}^\prime \prec \mathbf{G} \}$, surrogate training data $D_0 \leftarrow \{ (\mathbf{G}, \mathcal{P}_\mathbf{G}) \mid \mathbf{G} \in P_0 \}$;
Let $k \leftarrow 0$;
\WHILE{evaluation budget is not exhausted}
\STATE Train the RF surrogate on $D_k$;
\STATE $Q_k \leftarrow$ Generate candidate solutions from $P_k$ using crossover and mutation operators;
\STATE Predict the performance of $Q_k$ using the surrogate;
\STATE $Q'_k \leftarrow$ Select predicted non-dominated candidate solutions from $Q_k$;
\FOR {each $\mathbf{G} \in Q'_k$}
    \STATE $\mathcal{J}(\mathbf{G}) \leftarrow$ Generate a modular expert set according to Eq.~\eqref{eq: Modular_expert_generation};
    \STATE Evaluate $\mathbf{G}$ using $\mathcal{J}(\mathbf{G})$ via Eq.~\eqref{eq: performance} and Eq.~\eqref{eq: cost};
\ENDFOR
\STATE $P_{k+1} \leftarrow$ Select the top individuals from $P_{k} \cup Q'_k$;
\STATE $\mathcal{G}^\ast \leftarrow \{ \mathbf{G} \in (\mathcal{G}^\ast \cup Q'_k) \mid \nexists \mathbf{G}^\prime \in (\mathcal{G}^\ast \cup Q'_k)$ such that $\mathbf{G}^\prime \prec \mathbf{G} \}$;
\STATE $D_{k+1} \leftarrow D_k \cup \{ (\mathbf{G}, \mathcal{P}_\mathbf{G}) \mid \mathbf{G} \in Q'_k \}$;
\STATE $k \leftarrow k+1$;
\ENDWHILE
\STATE Construct the reusable modular expert library $\mathbb{L} = \{ \mathcal{J}(\mathbf{G}^\ast) \mid \mathbf{G}^\ast \in \mathcal{G}^\ast \}$.
% \STATE \textbf{return} $\mathcal{G}^\ast$, $\mathbb{L}$
\end{algorithmic}
\end{algorithm}

The on-demand reuse stage aims to achieve input-aware adaptability through modular expert recombination, thereby significantly enhancing the reusability of high-quality merging configurations and improving inference efficiency. As detailed in Algorithm~\ref{alg: on-demand_reuse}, leveraging the Pareto-optimal merging configurations obtained from the optimization process, this stage first selects a configuration subject to the given storage constraint and retrieves the corresponding modular expert set from the expert library (lines 1-2). Subsequently, a mapping from task patterns to modular expert chains is established to facilitate rapid and efficient module recombination (line 3). At inference time, a trained lightweight routing network identifies the task pattern that best matches the specific input. Based on this prediction, modular experts are dynamically recombined following the mapped expert chain, efficiently assembling an input-specific merged model (lines 4-13). Notably, the mechanism of reusing offline-derived modules naturally supports batch processing, as similar inputs tend to activate the same modular experts.

\begin{algorithm}[tb]
\caption{On-Demand Reuse in MERGE}
\label{alg: on-demand_reuse}
\textbf{Input}: Pareto-optimal set $\mathcal{G}^\ast$, modular expert library $\mathbb{L}$, routing network $\mathcal{R}$, storage constraint $\pi$, input data $\boldsymbol{X}$\\
\textbf{Output}: Output $\boldsymbol{Y}$ for input $\boldsymbol{X}$
\begin{algorithmic}[1] %[1] enables line numbers
\STATE $\mathbf{G}_{\pi} \leftarrow$ Select a configuration from $\mathcal{G}^\ast$ based on $\pi$;
\STATE $\mathcal{J}(\mathbf{G}_\pi) \leftarrow$ Retrieve the modular expert set from $\mathbb{L}$; 
\STATE $\mathcal{M}_{\mathbf{G}_{\pi}} \leftarrow$ Establish a mapping from task patterns to expert chains;
\STATE $\boldsymbol{Y} \leftarrow \emptyset$;
\FOR{$\boldsymbol{x} \in \boldsymbol{X}$}
    \STATE $\hat{t} \leftarrow$ Predict the optimal task pattern via Eq.~\eqref{eq: predict_task_pattern};
    \STATE $\boldsymbol{X}_{\hat{t}} \leftarrow$ Assign the instance $\boldsymbol{x}$ to the group;
\ENDFOR
\FOR{each non-empty $\boldsymbol{X}_{\hat{t}}$}
    \STATE $\mathbf{g}_{\hat{t}} \leftarrow \mathcal{M}_{\mathbf{G}_{\pi}}(\hat{t})$;
    \STATE $\widetilde{\theta}_{\hat{t}} \leftarrow$ Construct a specialized model by recombining modular experts via Eq.~\eqref{eq: model_reconstruction};
    \STATE $\boldsymbol{Y} \leftarrow \boldsymbol{Y} \cup f(\boldsymbol{X}_{\hat{t}}; \widetilde{\theta}_{\hat{t}})$;
\ENDFOR
% \STATE \textbf{return} $Y$
\end{algorithmic}
\end{algorithm}

\subsection{Component-Wise Merging Configuration Representation}
\label{subsec:Fine-grained Solution Representation}
For clarity, we restate the key notations from Section~\ref{sec:Preliminary and Related Work} as follows: $\theta_{\text{pre}}$ and $\{\theta_t\}_{t=1}^T$ denote the parameters of the pre-trained model and the task-specific models, respectively. The task vector for task $t$ is defined as $\tau_t = \theta_t - \theta_{\text{pre}}$.
From a modular perspective, these pre-trained and task-specific models can be decomposed into distinct functional components. Accordingly, their parameters are represented as $ \theta_{\text{pre}} = \{\theta_{\text{pre}}^l\}_{l=1}^{L}$ and $\theta_t = \{\theta_t^l\}_{l=1}^{L}$, where $L \in \mathbb{N}^+$ denotes the number of model components. 

To avoid the detrimental effects of indiscriminate merging, it is necessary to identify homologous components beneficial for merging and segregate those prone to catastrophic conflicts. Thus, a component-wise merging configuration is represented by a grouping matrix $\mathbf{G} = [ g_{t,l} ] \in \mathbb{N}^{T \times L}$, where each element $g_{t,l} \in \{1,2,\dots,N_l\}$ specifies the group assignment for component $l$ of task-specific model $t$. Here, $N_l$ denotes the number of distinct groups for component $l$, with $1 \leq N_l \leq T$. For each group $n \in \{1, 2, \ldots, N_l\}$, the associated collection of parameter subsets is defined as $\Theta_l^n = \{ \theta_t^l \mid g_{t,l} = n \}$. 

To enhance storage efficiency in fine-grained merging, we introduce an intra-group merging function $\mathcal{F}$ and a conditional compression function $\mathcal{Q}$, which are applied to form each modular expert $\hat{\theta}_l^n$:
\begin{equation}
\label{eq: Modular_expert_generation}
\hat{\theta}_l^n = \mathcal{Q} \left( \mathcal{F}\left( \theta_{\text{pre}}^l,\ \Theta_l^n \right) \right).
\end{equation}
Accordingly, the merging configuration $\mathbf{G}$ guides the construction of a reusable modular expert set, denoted by $\mathcal{J}(\mathbf{G}) = \{ \hat{\theta}_l^n \mid l \in \{1,2,\dots, L\}, n \in \{1,2,\dots,N_l\} \}$.

\subsubsection{Intra-Group Merging} 

When $|\Theta_l^n| \geq 2$, components from specific models are merged into a shared module using an arbitrary static merging function $\mathcal{F}$. For example, using Task Arithmetic \cite{ilharco2022editing} as the base method, the merged module is computed as:
\begin{equation}
\label{eq: intra-group_merging}
\mathcal{F}\left( \theta_{\text{pre}}^l,\ \Theta_l^n \right) = \theta_{\text{pre}}^l + \lambda \sum_{\theta_t^l \in \Theta_l^n} (\theta_t^l - \theta_{\text{pre}}^l)
\end{equation}
where $\lambda$ is a merging coefficient. Our MERGE method is compatible with various merging techniques and reduces storage cost by enabling intra-group parameter sharing.

\subsubsection{Conditional Compression} 

Parameter merging reshapes the parameter distribution \cite{wortsman2022model}. Merging functionally aligned modules typically induces dynamic range contraction, thus reducing the expected quantization error \cite{gong2025survey}. This phenomenon provides an endogenous signal to identify quantization-friendly units, further enhancing storage efficiency \cite{kuzmin2023pruning}.

For each quantization unit $u \in \{1,2,\dots,U_l\}$ (e.g., per-tensor or per-channel) in a given module, we apply asymmetric uniform quantization \cite{li2024contemporary} only if its dynamic range is less than or equal to the average dynamic range of the corresponding units from the source parameter subsets in $\Theta_l^n$. If this criterion is satisfied, each full-precision parameter $w$ within the specific unit is projected to the nearest value in the discrete grid $\mathcal{B} = \{ s(q - z) \mid q \in \mathbb{Z},\ q_{\min} \leq q \leq q_{\max} \}$. Here, $[q_{\min}, q_{\max}]$ denotes the integer range for $b$-bit quantization, and $s$ and $z$ are unit-specific scale and zero-point, respectively. The final projection is expressed as:
\begin{equation}
\label{eq: uniform_quantize}
\mathcal{Q}(w) = \arg\min_{\varpi \in \mathcal{B}} |w - \varpi|.
\end{equation}

\subsection{Pareto-Optimal Merging Configuration Search}
\label{subsec:Pareto-Optimal Solution Search}
Achieving component-wise fine-grained merging necessitates effective management of the trade-off between performance and storage efficiency. Specifically, maintaining a large collection of high-fidelity expert modules improves cross-task adaptability and performance, but also incurs substantial storage overhead. Moreover, the vast and discrete search space renders exhaustive evaluation of all possible component-wise merging configurations computationally infeasible. Accordingly, we formulate the component-wise merging process as a bi-objective optimization problem and develop a surrogate-assisted evolutionary algorithm to efficiently identify a set of Pareto-optimal merging configurations.

\subsubsection{Bi-Objective Problem Formulation}
We formulate the search for optimal component-wise merging configurations as a bi-objective optimization problem that simultaneously minimizes storage cost and maximizes cross-task performance.

The first objective evaluates the average cross-task performance achieved by a merging configuration $\mathbf{G}$: 
\begin{equation}
\label{eq: performance}
\mathcal{P}_\mathbf{G} = \frac{1}{T} \sum_{t=1}^{T} \left( \frac{1}{|\mathcal{D}_t^\prime|} \sum\nolimits_{(x_t,y_t) \in \mathcal{D}_t^\prime} \Phi_t \left( f_{\widetilde{\theta}_t}(x_t), y_t \right) \right).
\end{equation}
Here, $\widetilde{\theta}_{t} = \bigoplus_{l=1}^L\hat{\theta}_l^{g_{t,l}}$ is constructed by recombining modular experts from $\mathcal{J}(\mathbf{G})$. For each task, performance is evaluated using the task-specific evaluation protocol $\Phi_t(\cdot)$ on a size-constrained validation set $\mathcal{D}_t^\prime$. Notably, we minimize $-\mathcal{P}_\mathbf{G}$ to ensure both objectives are minimized.

The second objective is storage cost, which quantifies the storage requirements of the modular experts generated by configuration $\mathbf{G}$. This cost is normalized by the size of the full-precision pre-trained model to provide a standardized measure:

\begin{equation}
\label{eq: cost}
    \mathcal{C}_\mathbf{G} = \sum_{l=1}^{L} \sum_{n=1}^{N_l} \sum_{u=1}^{U_l} \left(\frac{b_l^{n,u} \cdot d_l^{n,u}}{b^+ \cdot d} + \frac{2 \cdot \mathbb{I}(b_l^{n,u} < b^+)}{d}\right).
\end{equation}
The first term represents the storage demand for model parameters, while the second term accounts for quantization metadata. Here, $d$ and $d_l^{n,u}$ denote the parameter counts of the full model and the specific unit, respectively, $b^+$ and $b_l^{n,u}$ represent their corresponding bit-widths, and $\mathbb{I}(\cdot)$ is the indicator function. 

\subsubsection{Surrogate-Assisted Optimization}

Our goal is to identify the Pareto-optimal set $\mathcal{G}^\ast \subseteq \mathcal{G}$, which consists of non-dominated solutions $\mathcal{G}^\ast = \{ \mathbf{G} \in \mathcal{G} \mid \nexists \mathbf{G}^\prime \in \mathcal{G} \text{ such that } \mathbf{G}^\prime \prec \mathbf{G} \}$. For each Pareto-optimal solution, no other solution in $\mathcal{G}$ can strictly improve one objective without degrading the other.

To efficiently explore the vast, discrete search space and reduce the cost of expensive evaluations, we introduce a surrogate-assisted bi-objective evolutionary algorithm. Specifically, we employ Random Forest (RF) \cite{breiman2001random} as the surrogate to approximate the performance function, as its tree-based structure is well-suited for high-dimensional discrete spaces.

The optimization process begins by initializing a population $P_0$ using K-means clustering \cite{hartigan1979algorithm}, which is applied to homologous components across task-specific models. These solutions are evaluated to form the initial surrogate training set, $D_0 = \{ (\mathbf{G}, \mathcal{P}_\mathbf{G}) \mid \mathbf{G} \in P_0 \}$. In each subsequent generation $k$, candidate solutions $Q_k$ are generated from the current population $P_k$ using evolutionary operators, such as selection, crossover, and mutation \cite{deb2002fast}. The surrogate then rapidly predicts their performance. Based on these predictions and the exact storage cost, a promising subset of non-dominated solutions $Q'_k \subseteq Q_k$ is selected for real evaluation. Subsequently, the newly evaluated solutions are used to form the next-generation population $P_{k+1}$, update the current Pareto-optimal set $\mathcal{G}^\ast$, and augment the surrogate training data, $D_{k+1} = D_{k} \cup \{ (\mathbf{G}, \mathcal{P}_\mathbf{G}) \mid \mathbf{G} \in Q'_k \}$. This process iterates until a predefined evaluation budget is exhausted.

\subsection{On-Demand Modular Expert Recombination}
\label{subsec:On-Demand Modular Expert Recombination}
After obtaining the Pareto-optimal set $\mathcal{G}^\ast$ through offline optimization, we construct a modular expert library $\mathbb{L} = \{ \mathcal{J}(\mathbf{G}^\ast) \mid \mathbf{G}^\ast \in \mathcal{G}^\ast \}$, where each $\mathcal{J}(\mathbf{G}^\ast)$ denotes a set of modular experts generated from a specific merging configuration in the Pareto-optimal set. At inference time, users select a configuration $\mathbf{G}_\pi \in \mathcal{G}^\ast$ that satisfies their storage constraints and deploy the corresponding modular expert set $\mathcal{J}(\mathbf{G}_\pi)$. 

To enable input-aware adaptability, MERGE leverages the selected modular expert set to dynamically assemble specialized models. This process consists of two key steps: (1) mapping various task patterns to expert chains, which represent the recombination paths of appropriate modular experts, and (2) assembling the input-specific merged model by routing each input to the optimal task pattern and recombining the selected modular experts according to the determined expert chain. 

\subsubsection{Task Pattern to Expert Chain Mapping}
Given a deployed expert set $\mathcal{J}(\mathbf{G}_\pi)$, the merging configuration $\mathbf{G}_\pi$ defines a set of expert chains, where each chain specifies a sequence of modular experts to be activated for a particular task pattern, thereby enabling efficient module recombination. Formally, for each $t \in \mathcal{T}$, where $\mathcal{T}$ denotes the set of all task patterns, the corresponding expert chain is represented as $\mathbf{g}_t = (g_{t,1}, g_{t,2}, \ldots, g_{t,L}) \in \prod_{l=1}^{L} \{1,2, \ldots, N_l\}$, where $g_{t,l}$ indicates the group index of the modular expert selected for model component $l$. 

Accordingly, the mapping $\mathcal{M}_{\mathbf{G}_{\pi}}: \mathcal{T} \to \mathcal{H}_{\mathbf{G}_{\pi}}$, associates each task pattern with an expert chain, where $\mathcal{H}_{\mathbf{G}_{\pi}}$ denotes the set of all expert chains defined by $\mathbf{G}_\pi$. This mapping supports both one-to-one and many-to-one associations depending on the merging configuration. Specifically, in the one-to-one case, distinct task patterns are mapped to distinct chains, i.e., $\forall t_i \neq t_j,\ \mathbf{g}_{t_i} \neq \mathbf{g}_{t_j}$, and in the many-to-one case, a group of different task patterns shares an identical expert chain, i.e., there exists a subset $\mathbb{T} \subseteq \mathcal{T}$ with $|\mathbb{T}| \geq 2$, such that $\forall t_i, t_j \in \mathbb{T},\ \mathbf{g}_{t_i} = \mathbf{g}_{t_j}$. 

\subsubsection{Input-Specific Model Assembly}

We train a lightweight routing network $\mathcal{R}$, implemented as a three-layer fully connected neural network, to guide the assembly of input-specific models. For a given input $\boldsymbol{x}$, the routing network predicts the optimal task pattern $\hat{t}$ based on high-dimensional features $\mathbf{h}(\boldsymbol{x})$ extracted from the pre-trained model: 
\begin{equation}
\label{eq: predict_task_pattern}
\hat{t} = \arg\max\limits_{t \in \mathcal{T}} \left( \mathcal{R}(\mathbf{h}(\boldsymbol{x}), t) \right).
\end{equation}
The corresponding expert chain $\mathbf{g}_{\hat{t}}$ is then obtained from the mapping $\mathcal{M}_{\mathbf{G}_{\pi}}$, which is used to activate the required modular experts from the deployed set $\mathcal{J}(\mathbf{G}_{\pi})$. These activated modular experts are subsequently assembled to form a specialized model for the current input:
\begin{equation}
\label{eq: model_reconstruction}
    \widetilde{\theta}_{\hat{t}} = \bigoplus_{l=1}^L 
    \Gamma(\mathcal{J}(\mathbf{G}_\pi), l, g_{\hat{t},l}).
\end{equation}
Here, $\Gamma(\mathcal{J}(\cdot),l,n)$ is a selector function that retrieves the appropriate modular expert $\hat{\theta}_l^n $ from $\mathcal{J}(\cdot)$, corresponding to component $l$ and group $n$, where $n$ is determined by the assigned group index $g_{\hat{t},l} \in \mathbf{g}_{\hat{t}}$. 

A key advantage of MERGE is its natural support for efficient inference compared to instance-specific merging and inference in existing dynamic merging approaches. By clustering inputs whose predicted patterns are mapped to an identical expert chain, all instances within the same group can be processed together, as they recombine the same modular experts and share the same constructed model.

\section{Experiments}
\label{sec:Experiments}
In this section, we evaluate the effectiveness of MERGE through extensive experiments. We first introduce the models, datasets, and evaluation protocols for three merging scenarios in Section~\ref{subsec:Models, Datasets, and Evaluation Protocols}, followed by the baselines and implementation details in Sections~\ref{subsec:Baselines} and \ref{subsec:Implementation Details}. Next, we present comparative results across different scenarios in Section~\ref{subsec:Experimental Results}. Section~\ref{subsec:MERGE Analysis} provides exploratory experiments that investigate MERGE's generalization, merging utility, and component-wise model integration behavior. Finally, ablation studies quantify the impact of MERGE's key strategies in Section~\ref{subsec:Ablation Study}.

\subsection{Models, Datasets, and Evaluation Protocols}
\label{subsec:Models, Datasets, and Evaluation Protocols}
\subsubsection{Merging Vision Models}
Following the settings in \cite{ilharco2022editing}, we adopt ViT-B/32 and ViT-L/14 \cite{radford2021learning} as pre-trained models, each independently fine-tuned on eight image classification datasets. These vision tasks include SUN397 \cite{xiao2010sun} for scene classification, Cars \cite{krause20133d} for car classification, RESISC45 \cite{cheng2017remote} for remote sensing scene classification, EuroSAT \cite{helber2019eurosat} for satellite classification, SVHN \cite{netzer2011reading} for real-world digit classification, GTSRB \cite{stallkamp2011german} for traffic sign classification, MNIST \cite{lecun1998mnist} for handwritten digit recognition, and DTD \cite{cimpoi2014describing} for texture classification. The performance is measured using accuracy. 

\subsubsection{Merging FFT Language Models}
RoBERTa-base \cite{liu2021robustly} and GPT-2 \cite{radford2019language} serve as pre-trained models. Following \cite{huang2024emr}, RoBERTa-base is fine-tuned on eight GLUE datasets \cite{wang2018glue}, which are categorized into three types. First, single-sentence tasks include CoLA \cite{warstadt2019neural} for linguistic acceptability classification, and SST-2 \cite{socher2013recursive} for sentiment analysis. Second, similarity and paraphrase tasks include MRPC \cite{dolan2005automatically} for paraphrase recognition, STS-B \cite{cer2017semeval} for semantic textual similarity assessment, and QQP \cite{quora2017} for question pair similarity evaluation. Finally, inference tasks include MNLI \cite{williams2018broad} for natural language inference, QNLI \cite{rajpurkar2016squad} for question-answering inference, and RTE \cite{giampiccolo2007third} for textual entailment recognition. The setup for GPT-2 is analogous, except that the STS-B task is excluded. Evaluation metrics include the Matthews correlation coefficient for CoLA, the average Pearson and Spearman correlation coefficients for STS-B, and accuracy for the other tasks \cite{wang2018glue}.

\subsubsection{Merging PEFT Models}
Following \cite{yadav2023ties}, we fine-tune the pre-trained T0-3B model \cite{sanh2021multitask} on eleven datasets using the (IA)$^3$ \cite{liu2022few} PEFT method. These tasks include RTE \cite{giampiccolo2007third} for textual entailment recognition, CB \cite{de2019commitmentbank} for commitment semantics understanding, Winogrande \cite{sakaguchi2021winogrande} for commonsense reasoning, WiC \cite{pilehvar2019wic} for word sense disambiguation, WSC \cite{levesque2012winograd} for pronoun resolution, COPA \cite{roemmele2011choice} for causal reasoning, H-SWAG \cite{zellers2019hellaswag} for commonsense natural language inference, Story Cloze \cite{sharma2018tackling} for story completion judgment, and ANLI \cite{nie2020adversarial} from R1 to R3 for adversarial natural language inference. Accuracy is used as the evaluation metric for all tasks.

\subsection{Baselines}
\label{subsec:Baselines}
We compare our proposed MERGE method with three categories of baseline approaches.
\subsubsection{Non-Merging Methods}
\begin{itemize}
    \item Individual models are task-specific models fine-tuned for specific tasks using either FFT or PEFT. 
    \item Traditional MTL jointly trains a single model on labeled data from all tasks.
    \item Pre-trained models are general-purpose models trained on large-scale datasets.
\end{itemize}

\subsubsection{Static Model Merging Methods}
\begin{itemize}
    \item Weight Averaging \cite{wortsman2022model} refers to simply averaging model parameters to obtain the merged model. 
    \item Fisher Merging \cite{matena2022merging} estimates the importance of each parameter in individual models using Fisher information matrices, and merges models accordingly. 
    \item RegMean \cite{jin2022dataless} provides a closed-form solution to model merging, aiming to derive a merged model whose outputs closely match those of the individual models. Inner-product matrices are required for merging. 
    \item Task Arithmetic \cite{ilharco2022editing} aggregates task vectors with a global merging coefficient to construct a merged model. 
    \item Ties-Merging \cite{yadav2023ties} addresses conflicts among multiple models by eliminating redundant parameters and resolving symbol conflicts. 
    \item  Breadcrumbs \cite{davari2024model} masks out both low-magnitude values and outliers to mitigate task conflicts. 
    \item PCB-Merging \cite{du2024parameter} ranks parameters by importance via a parameter competition balancing strategy, dropping less important parameters while rescaling critical ones.
    \item AdaMerging \cite{yang2023adamerging} extends Task Arithmetic by introducing an unsupervised approach to learn per-task and per-layer merging coefficients.
\end{itemize}

\subsubsection{Dynamic Model Merging Methods}
\begin{itemize}
    \item Twin-Merging \cite{lu2024twin} builds a shared expert via static merging and compresses task-specific experts with singular value decomposition (SVD). At inference, a routing network weights and merges task-specific and shared experts to generate instance-specific models.
    \item EMR-Merging \cite{huang2024emr} elects a unified model and retains a mask and rescaler for each task. At inference, task-specific parameters are rescaled and combined with the unified model. 
\end{itemize}

\begin{table*}[htbp]
% \small
\footnotesize
\caption{Comparison of normalized storage cost $\mathcal{C}$ and performance $\mathcal{P}$ across experiments. For each MERGE variant, three representative solutions from its Pareto front are selected: $G_1$, the storage-minimal solution outperforming the corresponding base method; $G_2$, the storage-minimal solution exceeding the best merging baseline; and $G_3$, the performance-maximal solution.}
\scriptsize
\centering
\setlength{\tabcolsep}{2.5pt}
\begin{tabular}{l|cc|cc|cc|cc|cc}
\toprule
\multirow{3}{*}{Methods} 
    & \multicolumn{4}{c|}{Vision Models} 
    & \multicolumn{4}{c|}{Fully Fine-Tuned Language Models} 
    & \multicolumn{2}{c}{PEFT Language Models} \\
\cmidrule(lr){2-5} \cmidrule(lr){6-9} \cmidrule(lr){10-11}
    & \multicolumn{2}{c|}{ViT-B/32 on 8 tasks} 
    & \multicolumn{2}{c|}{ViT-L/14 on 8 tasks} 
    & \multicolumn{2}{c|}{RoBERTa on 8 tasks} 
    & \multicolumn{2}{c|}{GPT-2 on 7 tasks} 
    & \multicolumn{2}{c}{(IA)$^3$ on 11 tasks} \\
\cmidrule(lr){2-3} \cmidrule(lr){4-5} \cmidrule(lr){6-7} \cmidrule(lr){8-9} \cmidrule(lr){10-11}
    & $\mathcal{C}$ ($\downarrow$) & $\mathcal{P}$ ($\uparrow$)
    & $\mathcal{C}$ ($\downarrow$) & $\mathcal{P}$ ($\uparrow$)
    & $\mathcal{C}$ ($\downarrow$) & $\mathcal{P}$ ($\uparrow$)
    & $\mathcal{C}$ ($\downarrow$) & $\mathcal{P}$ ($\uparrow$)
    & $\mathcal{C}$ ($\downarrow$) & $\mathcal{P}$ ($\uparrow$) \\
\midrule
Individual & 8.00 & 90.5 & 8.00 & 94.2 & 8.00 & 85.6 & 7.00 & 82.0 & 1.00 & 71.4 \\
Traditional MTL & 1.00 & 89.1 & 1.00 & 93.1 & - & - & - & - & 1.00 & 73.1 \\
Pre-trained & 1.00 & 48.3 & 1.00 & 65.2 & 1.00 & 31.5 & 1.00 & 44.5 & 1.00 & 53.1 \\
\midrule
Weight Averaging & 1.00 & 65.9 & 1.00 & 79.5 & 1.00 & 56.4 & 1.00 & 56.1 & 1.00 & 58.0 \\
Fisher Merging & 1.00 & 68.3 & 1.00 & 82.2 & 1.00 & 58.7 & 1.00 & 58.7 & 1.00 & 62.2 \\
RegMean & 1.00 & 71.8 & 1.00 & 83.7 & 1.00 & 70.0 & 1.00 & 68.8 & 1.00 & 58.0 \\
Task Arithmetic & 1.00 & 60.5 & 1.00 & 83.1 & 1.00 & 60.4 & 1.00 & 69.7 & 1.00 & 59.2 \\
Ties-Merging & 1.00 & 72.4 & 1.00 & 85.9 & 1.00 & 63.5 & 1.00 & 64.8 & 1.00 & 64.9 \\
Breadcrumbs & 1.00 & 73.2 & 1.00 & 85.3 & 1.00 & 68.2 & 1.00 & 67.5 & 1.00 & 58.4 \\
PCB-Merging & 1.00 & 76.4 & 1.00 & 86.8 & 1.00 & 67.9 & 1.00 & 69.7 & 1.00 & 66.0 \\
AdaMerging & 1.00 & 80.1 & 1.00 & 90.8 & - & - & - & - & - & - \\
\midrule
Twin-Merging & 1.16 & 87.8 & 1.16 & 92.7 & 1.16 & 78.3 & 1.14 & 80.6 & 1.00 & 66.8 \\
EMR-Merging & 1.25 & 88.8 & 1.25 & 93.7 & 1.25 & 74.2 & 1.22 & 80.4 & 1.00 & 67.1 \\
\midrule
\textbf{MERGE-WA ($G_1$)} & \textbf{0.3141$\pm$0.0} & 70.7$\pm$1.2 & \textbf{0.2818$\pm$0.0} & 79.9$\pm$0.0 & \textbf{0.3446$\pm$0.0} & 62.5$\pm$1.1 & \textbf{0.3080$\pm$0.0} & 63.5$\pm$1.4 & \textbf{0.2502$\pm$0.0} & 59.6$\pm$0.3 \\
\textbf{MERGE-WA ($G_2$)} & 1.2297$\pm$0.1 & 89.2$\pm$0.2 & 1.5137$\pm$0.1 & 93.9$\pm$0.1 & 0.8435$\pm$0.1 & 79.0$\pm$0.1 & 1.2593$\pm$0.1 & 81.0$\pm$0.2 & 0.2508$\pm$0.0 & 67.7$\pm$0.4 \\
\textbf{MERGE-WA ($G_3$)} & 1.7079$\pm$0.1 & \textbf{90.2$\pm$0.2} & 1.6142$\pm$0.2 & \textbf{94.0$\pm$0.1} & 1.6677$\pm$0.1 & \textbf{83.6$\pm$0.3} & 1.4892$\pm$0.2 & \textbf{81.3$\pm$0.4} & 0.2515$\pm$0.0 & \textbf{69.4$\pm$0.3} \\
\midrule
\textbf{MERGE-TA ($G_1$)} & \textbf{0.3141$\pm$0.0} & 64.2$\pm$0.4 & \textbf{0.2514$\pm$0.0} & 83.1$\pm$0.0 & \textbf{0.2875$\pm$0.0} & 60.8$\pm$0.0 & \textbf{0.2508$\pm$0.0} & 69.9$\pm$0.1 & \textbf{0.2502$\pm$0.0} & 63.4$\pm$0.1 \\
\textbf{MERGE-TA ($G_2$)} & 1.2554$\pm$0.0 & 89.2$\pm$0.2 & 1.5015$\pm$0.1 & 93.8$\pm$0.1 & 0.7322$\pm$0.0 & 78.7$\pm$0.1 & 1.2320$\pm$0.1 & 80.8$\pm$0.1 & 0.2505$\pm$0.0 & 68.6$\pm$0.6 \\
\textbf{MERGE-TA ($G_3$)} & 1.7451$\pm$0.1 & \textbf{90.3$\pm$0.1} & 1.7824$\pm$0.1 & \textbf{94.0$\pm$0.1} & 1.8544$\pm$0.2 & \textbf{83.7$\pm$0.1} & 1.5894$\pm$0.1 & \textbf{81.6$\pm$0.1} & 0.2506$\pm$0.0 & \textbf{69.1$\pm$0.1} \\
\midrule
\textbf{MERGE-TM ($G_1$)} & \textbf{0.2513$\pm$0.0} & 73.9$\pm$0.0 & \textbf{0.2514$\pm$0.0} & 86.0$\pm$0.0 & \textbf{0.3446$\pm$0.0} & 67.2$\pm$1.2 & \textbf{0.3080$\pm$0.0} & 67.4$\pm$0.9 & \textbf{0.2502$\pm$0.0} & 65.5$\pm$0.5 \\
\textbf{MERGE-TM ($G_2$)} & 1.0162$\pm$0.0 & 89.2$\pm$0.2 & 1.3212$\pm$0.0 & 93.9$\pm$0.1 & 0.8277$\pm$0.1 & 79.4$\pm$0.9 & 1.1246$\pm$0.1 & 80.9$\pm$0.2 & 0.2506$\pm$0.0 & 67.8$\pm$0.4 \\
\textbf{MERGE-TM ($G_3$)} & 1.6324$\pm$0.2 & \textbf{90.2$\pm$0.1} & 1.4364$\pm$0.1 & \textbf{93.9$\pm$0.0} & 1.6748$\pm$0.2 & \textbf{83.5$\pm$0.3} & 1.5735$\pm$0.1 & \textbf{81.6$\pm$0.2} & 0.2511$\pm$0.0 & \textbf{69.4$\pm$0.5} \\
\midrule
\textbf{MERGE-BC ($G_1$)} & \textbf{0.3141$\pm$0.0} & 76.4$\pm$0.0 & \textbf{0.3148$\pm$0.0} & 85.6$\pm$0.5 & \textbf{0.3446$\pm$0.0} & 71.0$\pm$0.0 & \textbf{0.3080$\pm$0.0} & 71.1$\pm$0.0 & \textbf{0.2502$\pm$0.0} & 63.2$\pm$0.3 \\
\textbf{MERGE-BC ($G_2$)} & 1.2929$\pm$0.1 & 89.3$\pm$0.1 & 1.4843$\pm$0.3 & 93.9$\pm$0.1 & 0.7821$\pm$0.1 & 79.1$\pm$0.4 & 1.1659$\pm$0.2 & 80.8$\pm$0.1 & 0.2511$\pm$0.0 & 68.1$\pm$0.4 \\
\textbf{MERGE-BC ($G_3$)} & 1.8086$\pm$0.2 & \textbf{90.2$\pm$0.1} & 1.8093$\pm$0.2 & \textbf{94.0$\pm$0.1} & 1.9287$\pm$0.1 & \textbf{83.7$\pm$0.1} & 1.5779$\pm$0.1 & \textbf{81.4$\pm$0.2} & 0.2517$\pm$0.0 & \textbf{69.3$\pm$0.1} \\
\bottomrule
\end{tabular}
\label{tab:main_results}
\end{table*}

\begin{table}[htbp]
% \scriptsize
\tiny
% \footnotesize
\centering
\caption{Per-task results of merging ViT-B/32 models on 8 vision tasks.} 
\setlength{\tabcolsep}{2pt}
\begin{tabular}{l|cccccccc}
\toprule
\multirow{1}{*}{Methods} 
    & SUN397 & Cars & RESISC45 & EuroSAT & SVHN & GTSRB & MNIST & DTD \\
\midrule
Individual & 75.3 & 77.7 & 96.1 & 99.7 & 97.5 & 98.7 & 99.7 & 79.4 \\
Traditional MTL & 73.9 & 74.4 & 93.9 & 98.2 & 95.8 & 98.9 & 99.5 & 77.9 \\
Pre-trained & 63.2 & 59.6 & 60.2 & 46.8 & 31.6 & 32.6 & 48.3 & 44.4 \\
\midrule
Weight Averaging & 65.3 & 63.3 & 71.4 & 72.6 & 64.2 & 52.8 & 87.5 & 50.1 \\
Fisher Merging & 68.6 & 69.2 & 70.7 & 66.4 & 72.9 & 51.1 & 87.9 & 59.9 \\
RegMean & 65.3 & 63.5 & 75.6 & 78.6 & 78.1 & 67.4 & 93.7 & 52.0 \\
Task Arithmetic & 36.7 & 41.0 & 53.8 & 65.0 & 80.6 & 66.0 & 98.1 & 42.5 \\
Ties-Merging & 59.8 & 58.6 & 70.7 & 79.6 & 86.2 & 72.1 & 98.3 & 54.2 \\
Breadcrumbs & 65.6 & 62.9 & 75.0 & 78.6 & 79.0 & 71.0 & 96.7 & 57.0 \\
PCB-Merging & 66.7 & 65.5 & 78.5 & 79.3 & 86.4 & 77.1 & 98.2 & 59.1 \\
AdaMerging & 64.5 & 68.1 & 79.2 & 93.8 & 87.0 & 91.9 & 97.5 & 59.1 \\
\midrule
Twin-Merging & 73.6 & 71.7 & 92.1 & 99.3 & 95.3 & 97.2 & 99.1 & 74.0 \\
EMR-Merging & \textbf{75.2} & 72.8 & 93.5 & 99.5 & 96.9 & 98.1 & 99.6 & 74.4 \\
\midrule
\textbf{MERGE-WA ($G_1$)} & 65.2$\pm$0.1 & 62.8$\pm$0.2 & 72.5$\pm$0.5 & 75.2$\pm$0.9 & 87.0$\pm$4.9 & 63.3$\pm$12.5 & 88.5$\pm$4.2 & 51.0$\pm$0.4 \\
\textbf{MERGE-WA ($G_2$)} & 72.3$\pm$1.4 & 76.6$\pm$0.6 & 94.9$\pm$0.5 & 98.5$\pm$1.8 & 96.5$\pm$0.8 & 97.9$\pm$1.0 & 99.3$\pm$0.4 & 77.2$\pm$2.3 \\
\textbf{MERGE-WA ($G_3$)} & 75.0$\pm$0.1 & \textbf{77.1$\pm$0.2} & \textbf{95.9$\pm$0.0} & 99.3$\pm$0.8 & \textbf{97.4$\pm$0.1} & \textbf{98.2$\pm$0.4} & \textbf{99.6$\pm$0.2} & \textbf{79.1$\pm$0.1} \\
\midrule
\textbf{MERGE-TA ($G_1$)} & 38.6$\pm$0.9 & 41.5$\pm$0.9 & 60.8$\pm$7.6 & 68.1$\pm$2.3 & 80.5$\pm$1.7 & 81.5$\pm$8.8 & 98.1$\pm$0.1 & 44.7$\pm$0.9 \\
\textbf{MERGE-TA ($G_2$)} & 73.8$\pm$0.8 & 75.5$\pm$2.4 & 93.8$\pm$0.7 & 98.1$\pm$0.8 & 97.1$\pm$0.4 & 98.2$\pm$0.7 & 99.3$\pm$0.4 & 77.6$\pm$1.7 \\
\textbf{MERGE-TA ($G_3$)} & 74.8$\pm$0.2 & \textbf{77.2$\pm$0.0} & \textbf{95.7$\pm$0.2} & \textbf{99.7$\pm$0.0} & \textbf{97.4$\pm$0.1} & \textbf{98.7$\pm$0.1} & \textbf{99.7$\pm$0.0} & \textbf{79.2$\pm$0.3} \\
\midrule
\textbf{MERGE-TM ($G_1$)} & 62.5$\pm$0.0 & 60.3$\pm$0.0 & 74.6$\pm$0.0 & 77.6$\pm$0.0 & 86.9$\pm$0.0 & 74.1$\pm$0.0 & 98.5$\pm$0.0 & 56.3$\pm$0.0 \\
\textbf{MERGE-TM ($G_2$)} & 73.0$\pm$0.9 & 75.3$\pm$1.0 & 94.3$\pm$2.1 & 98.9$\pm$0.5 & 96.6$\pm$0.9 & 98.4$\pm$0.2 & 99.5$\pm$0.1 & 77.2$\pm$0.4 \\
\textbf{MERGE-TM ($G_3$)} & 74.9$\pm$0.2 & \textbf{77.3$\pm$0.4} & \textbf{95.6$\pm$0.2} & \textbf{99.7$\pm$0.1} & \textbf{97.4$\pm$0.1} & \textbf{98.6$\pm$0.2} & \textbf{99.6$\pm$0.1} & \textbf{78.8$\pm$0.4} \\
\midrule
\textbf{MERGE-BC ($G_1$)} & 67.0$\pm$0.0 & 63.3$\pm$0.0 & 77.5$\pm$0.0 & 75.9$\pm$0.0 & 78.3$\pm$0.0 & 94.9$\pm$0.0 & 96.8$\pm$0.0 & 57.4$\pm$0.0 \\
\textbf{MERGE-BC ($G_2$)} & 72.8$\pm$1.9 & 75.7$\pm$0.3 & 94.2$\pm$1.1 & 99.5$\pm$0.1 & 96.9$\pm$0.3 & 98.0$\pm$0.9 & 98.8$\pm$0.6 & 78.4$\pm$0.9 \\
\textbf{MERGE-BC ($G_3$)} & 74.8$\pm$0.2 & \textbf{76.6$\pm$0.7} & \textbf{95.8$\pm$0.1} & \textbf{99.7$\pm$0.1} & \textbf{97.3$\pm$0.2} & \textbf{98.7$\pm$0.1} & \textbf{99.7$\pm$0.0} & \textbf{79.2$\pm$0.1} \\
\bottomrule
\end{tabular}
\label{tab:ViTB32}
\end{table}

\subsection{Implementation Details} 
\label{subsec:Implementation Details}
We evaluate MERGE by adopting Weight Averaging (WA), Task Arithmetic (TA), Ties-Merging (TM), and Breadcrumbs (BC) for intra-group merging, thereby generating variants such as MERGE-WA. The original hyperparameters of these methods are retained, while the merging granularity is adjusted to align with our component-wise structure. The bit-width for conditional compression is set to eight. For surrogate-assisted bi-objective optimization, we adopt the NSGA-II algorithm \cite{deb2002fast}, with a population size of 20 and a maximum of 300 real evaluations per instance. Each evolutionary search is independently executed five times, and results are reported as the mean and standard deviation. Solution quality is measured using the hypervolume (HV) indicator \cite{audet2021performance}, which evaluates both convergence and diversity of Pareto-optimal solutions. Our lightweight routing network is trained on a dataset containing up to 1,000 instances per downstream task for 30 epochs with a learning rate of 5e-4. All merging experiments are performed on a single NVIDIA RTX A6000 GPU with 48~GB of RAM.

\begin{table}[htbp]
\tiny
\centering
\caption{Per-task results of merging ViT-L/14 models on 8 vision tasks.}
\setlength{\tabcolsep}{2pt}
\begin{tabular}{l|cccccccc}
\toprule
\multirow{1}{*}{Methods} 
    & SUN397 & Cars & RESISC45 & EuroSAT & SVHN & GTSRB & MNIST & DTD \\
\midrule
Individual & 82.3 & 92.4 & 97.4 & 100.0 & 98.1 & 99.2 & 99.7 & 84.1 \\
Traditional MTL & 80.8 & 90.6 & 96.3 & 96.3 & 97.6 & 99.1 & 99.6 & 84.4 \\
Pre-trained & 68.2 & 77.9 & 71.3 & 63.0 & 58.4 & 50.6 & 76.4 & 55.4 \\
\midrule
Weight Averaging & 72.1 & 81.6 & 82.6 & 91.3 & 78.2 & 70.6 & 97.0 & 62.8 \\
Fisher Merging & 69.2 & 88.6 & 87.5 & 93.5 & 80.6 & 74.8 & 93.3 & 70.0 \\
RegMean & 73.3 & 81.8 & 86.1 & 97.0 & 88.0 & 84.2 & 98.5 & 60.8 \\
Task Arithmetic & 72.5 & 79.2 & 84.5 & 89.6 & 89.2 & 86.5 & 99.1 & 64.3 \\
Ties-Merging & 76.5 & 85.0 & 89.3 & 94.9 & 90.3 & 83.3 & 99.0 & 68.8 \\
Model Breadcrumbs & 75.4 & 84.6 & 88.4 & 95.2 & 85.8 & 85.7 & 98.8 & 68.6 \\
PCB-Merging & 75.8 & 86.0 & 89.2 & 95.5 & 88.0 & 90.9 & 99.1 & 69.9 \\
AdaMerging & 79.0 & 90.3 & 90.8 & 96.2 & 93.4 & 98.0 & 99.0 & 79.9 \\
\midrule
Twin-Merging & 82.6 & 90.3 & 94.9 & 99.4 & 96.7 & 98.0 & 99.4 & 80.1 \\
EMR-Merging & \textbf{83.2} & 90.7 & 96.8 & 99.7 & 97.9 & 99.1 & 99.7 & 82.7 \\
\midrule
\textbf{MERGE-WA ($G_1$)} & 72.1$\pm$0.0 & 81.5$\pm$0.0 & 82.5$\pm$0.0 & 91.2$\pm$0.0 & 78.3$\pm$0.0 & 73.2$\pm$1.5 & 97.1$\pm$0.0 & 63.4$\pm$1.1 \\
\textbf{MERGE-WA ($G_2$)} & 81.9$\pm$0.1 & 92.2$\pm$0.2 & 97.4$\pm$0.0 & 99.6$\pm$0.2 & 98.0$\pm$0.2 & 98.7$\pm$0.6 & 99.6$\pm$0.0 & 83.5$\pm$0.7 \\
\textbf{MERGE-WA ($G_3$)} & 82.0$\pm$0.1 & \textbf{92.3$\pm$0.1} & \textbf{97.3$\pm$0.1} & 99.6$\pm$0.2 & \textbf{98.1$\pm$0.1} & 99.0$\pm$0.2 & \textbf{99.7$\pm$0.0} & \textbf{83.8$\pm$0.8} \\
\midrule
\textbf{MERGE-TA ($G_1$)} & 72.3$\pm$0.0 & 78.6$\pm$0.0 & 84.3$\pm$0.0 & 89.7$\pm$0.0 & 89.2$\pm$0.0 & 86.2$\pm$0.0 & 99.1$\pm$0.0 & 65.6$\pm$0.0 \\
\textbf{MERGE-TA ($G_2$)} & 81.8$\pm$0.2 & 92.0$\pm$0.3 & 97.1$\pm$0.2 & 99.5$\pm$0.3 & 98.0$\pm$0.1 & 98.8$\pm$0.3 & 99.7$\pm$0.1 & 83.8$\pm$0.4 \\
\textbf{MERGE-TA ($G_3$)} & 81.9$\pm$0.2 & \textbf{92.2$\pm$0.1} & \textbf{97.3$\pm$0.0} & \textbf{99.7$\pm$0.0} & \textbf{98.0$\pm$0.2} & \textbf{99.2$\pm$0.2} & \textbf{99.7$\pm$0.0} & \textbf{84.2$\pm$0.1} \\
\midrule
\textbf{MERGE-TM ($G_1$)} & 76.6$\pm$0.0 & 85.0$\pm$0.0 & 89.5$\pm$0.0 & 94.6$\pm$0.0 & 90.5$\pm$0.0 & 83.5$\pm$0.0 & 99.0$\pm$0.0 & 69.3$\pm$0.0 \\
\textbf{MERGE-TM ($G_2$)} & 82.1$\pm$0.1 & 92.2$\pm$0.2 & 97.2$\pm$0.3 & 99.6$\pm$0.1 & 98.0$\pm$0.1 & 98.7$\pm$0.5 & 99.7$\pm$0.0 & 83.3$\pm$0.3 \\
\textbf{MERGE-TM ($G_3$)} & 82.1$\pm$0.1 & \textbf{92.1$\pm$0.4} & \textbf{97.4$\pm$0.1} & \textbf{99.7$\pm$0.0} & \textbf{98.1$\pm$0.1} & \textbf{99.1$\pm$0.1} & \textbf{99.7$\pm$0.0} & \textbf{83.2$\pm$0.2} \\
\midrule
\textbf{MERGE-BC ($G_1$)} & 75.3$\pm$0.0 & 84.6$\pm$0.1 & 88.1$\pm$0.1 & 95.2$\pm$0.4 & 87.9$\pm$4.7 & 85.7$\pm$1.0 & 98.8$\pm$0.0 & 69.4$\pm$1.0 \\
\textbf{MERGE-BC ($G_2$)} & 81.7$\pm$0.2 & 91.9$\pm$0.3 & 97.0$\pm$0.1 & 99.7$\pm$0.1 & 97.8$\pm$0.2 & 98.9$\pm$0.3 & 99.5$\pm$0.2 & 84.2$\pm$0.1 \\
\textbf{MERGE-BC ($G_3$)} & 82.0$\pm$0.1 & \textbf{92.2$\pm$0.2} & \textbf{97.2$\pm$0.1} & \textbf{99.7$\pm$0.0} & \textbf{98.1$\pm$0.0} & \textbf{99.1$\pm$0.2} & \textbf{99.7$\pm$0.0} & \textbf{84.2$\pm$0.1} \\
\bottomrule
\end{tabular}
\label{tab:ViTL14}
\end{table}

\subsection{Experimental Results}
\label{subsec:Experimental Results}
We evaluate the effectiveness of MERGE across diverse experimental settings, including different model scales, task types, and fine-tuning strategies. Table~\ref{tab:main_results} summarizes the key results. For each MERGE variant, we report three representative solutions from its Pareto front: $G_1$ is the storage-minimal solution that outperforms the corresponding base method, $G_2$ is the storage-minimal solution that surpasses the best merging baseline, and $G_3$ is the performance-maximal solution. Task-level performance is further detailed in Table~\ref{tab:ViTB32} for ViT-B/32 models, Table~\ref{tab:ViTL14} for ViT-L/14 models, Table~\ref{tab:RoBERTa} for RoBERTa models, Table~\ref{tab:GPT-2} for GPT-2 models, and Table~\ref{tab:PEFT} for T0-3B PEFT models. In all tables, the best results for each MERGE variant among merging baselines are highlighted in bold. 

\begin{figure*}[t]
\centering
\subfloat[]{%
    \includegraphics[width=0.55\linewidth, height=4cm]{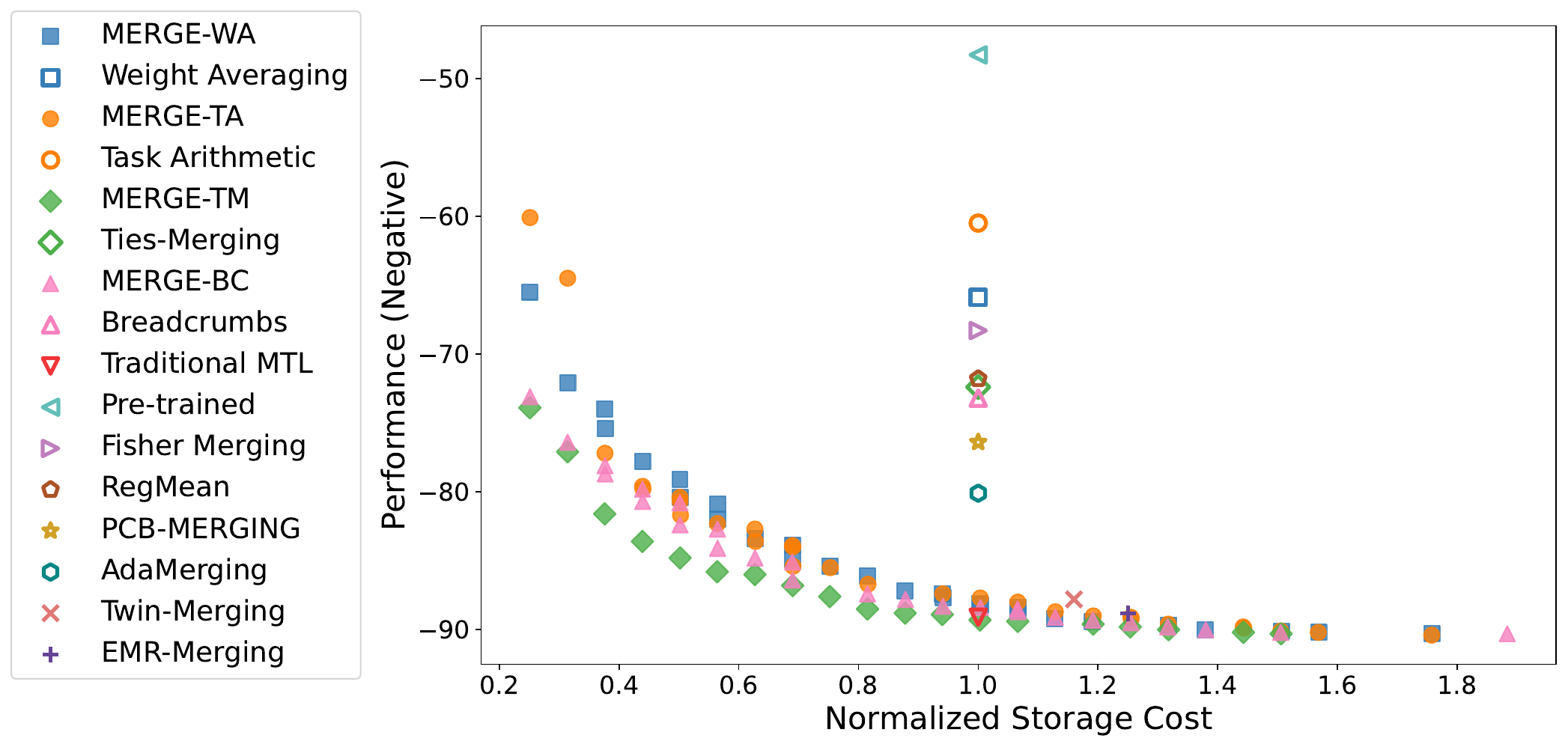}%
    \label{fig3:sub1}}
\hspace{0.01\textwidth}
\subfloat[]{%
    \includegraphics[width=0.4\linewidth, height=4cm]{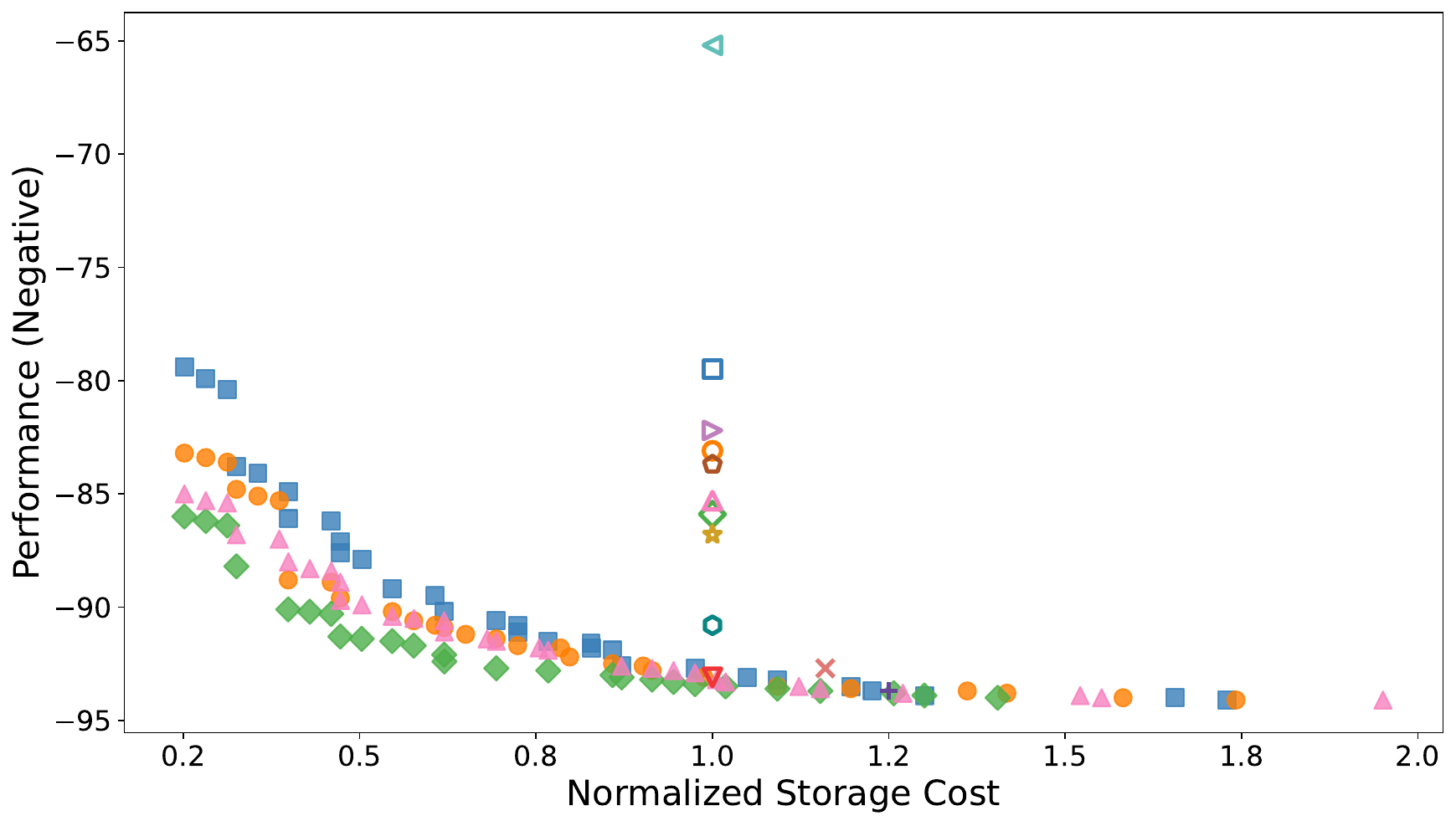}%
    \label{fig3:sub2}}

\subfloat[]{%
    \includegraphics[width=0.33\linewidth, height=4cm]{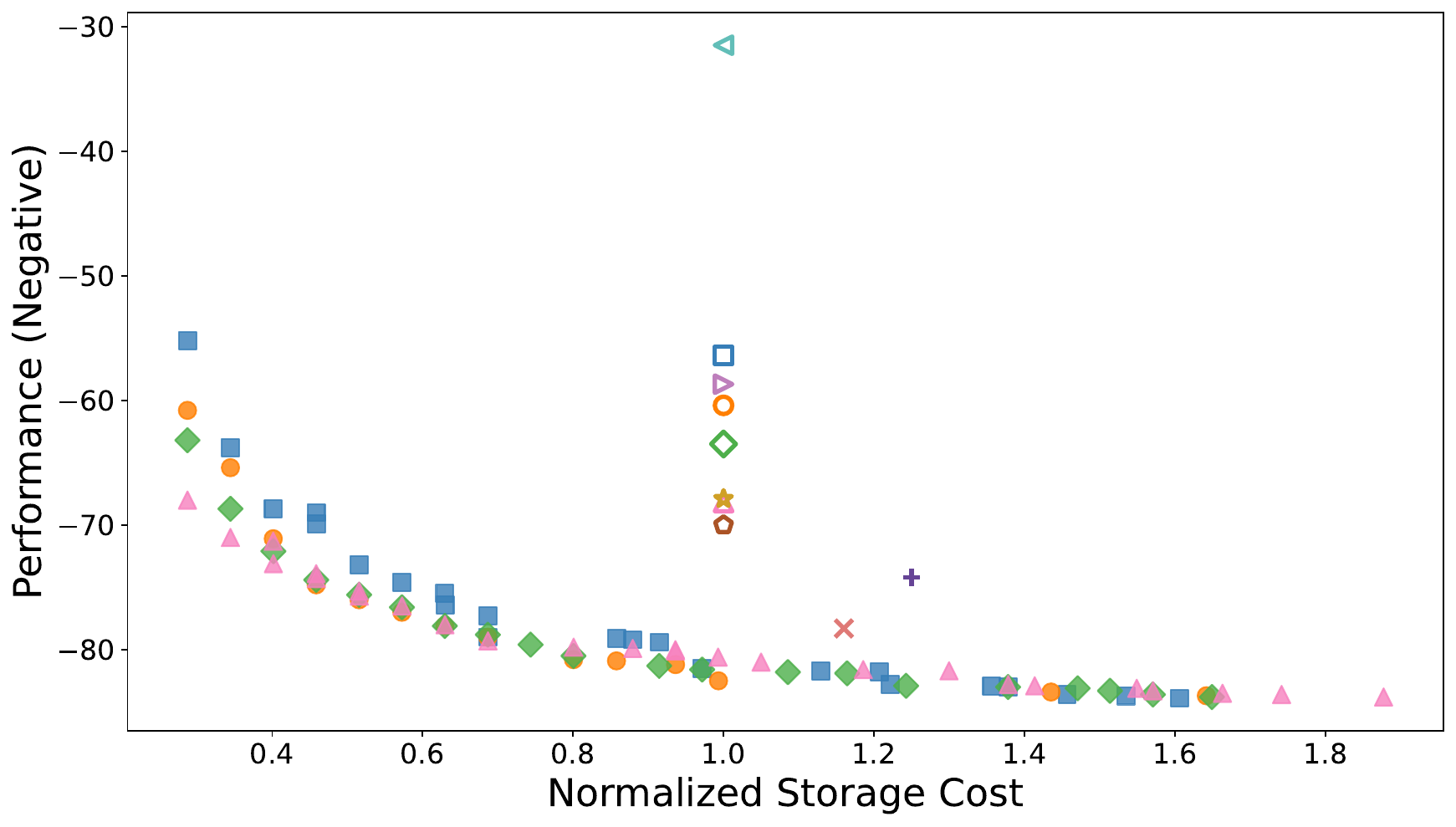}%
    \label{fig3:sub3}}
% \hspace{0.01\textwidth}
\subfloat[]{%
    \includegraphics[width=0.33\linewidth, height=4cm]{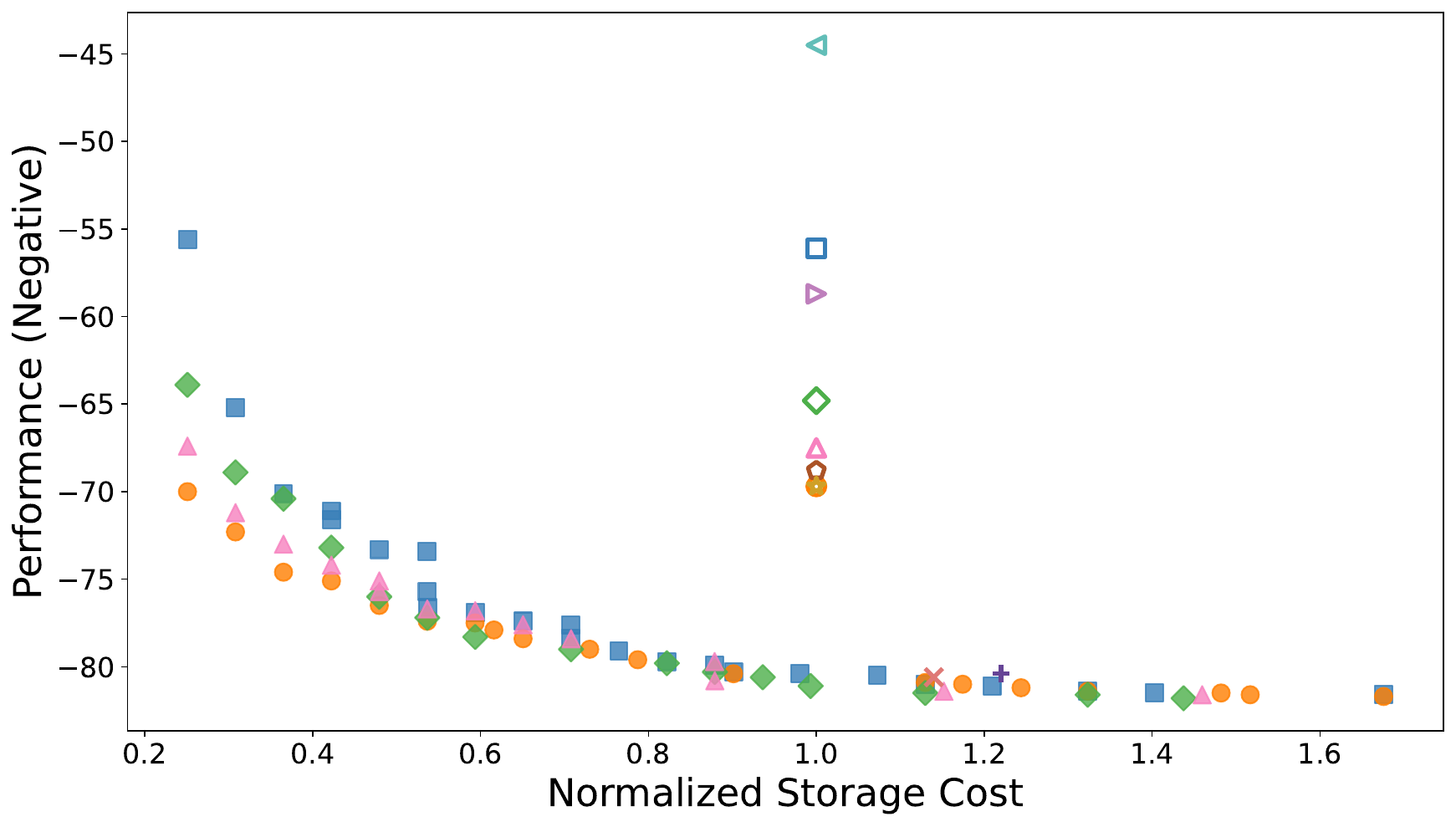}%
    \label{fig3:sub4}}
% \hspace{0.01\textwidth}
\subfloat[]{%
    \includegraphics[width=0.33\linewidth, height=4cm]{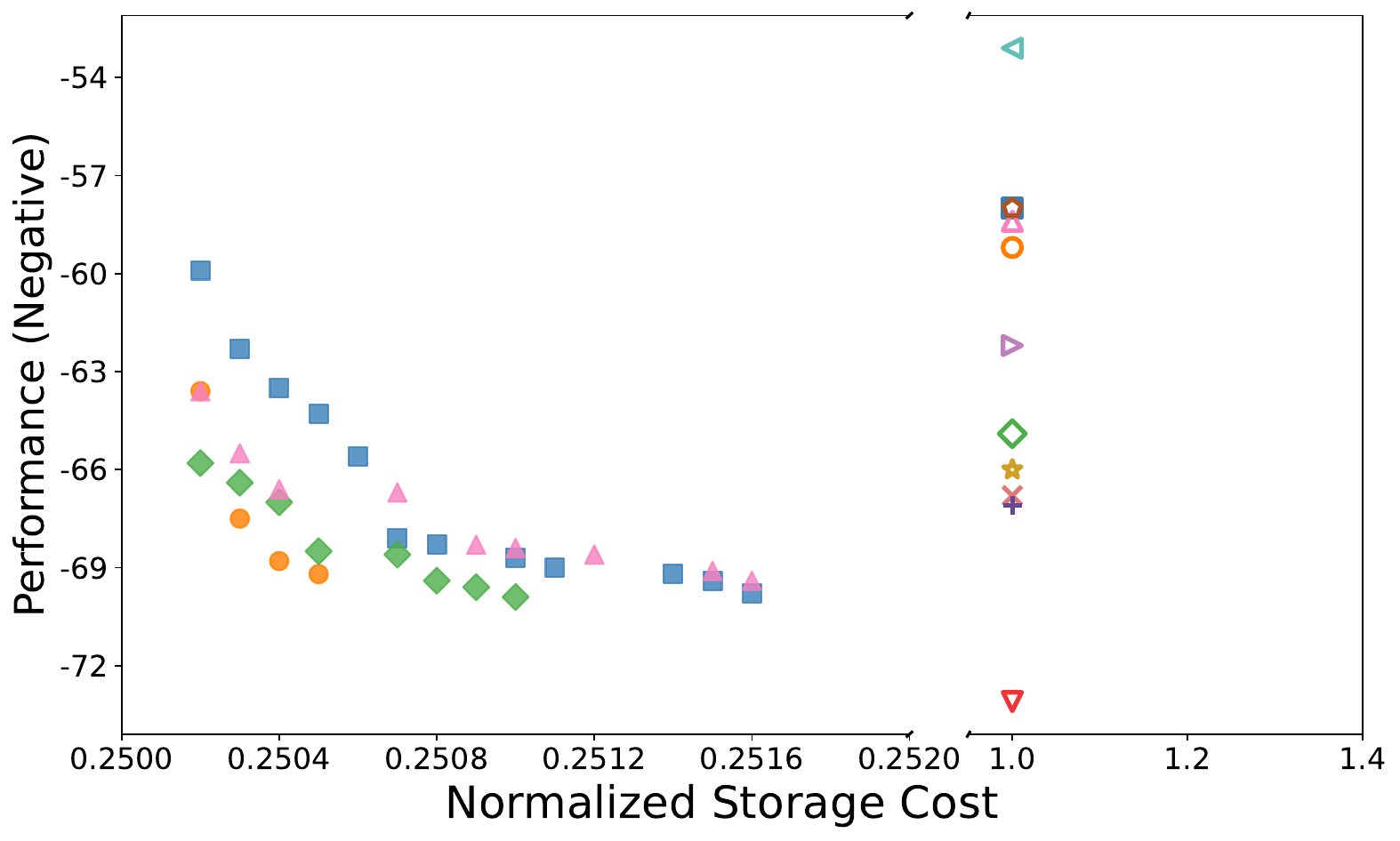}%
    \label{fig3:sub5}}
\caption{Comparison of Pareto-optimal solutions obtained by different methods across various models. (a) Merging ViT-B/32 models. (b) Merging ViT-L/14 models. (c) Merging RoBERTa models. (d) Merging GPT-2 models. (e) Merging PEFT models.}
\label{fig:all}
\end{figure*}

\begin{table}[htbp]
\tiny
\centering
\caption{Per-task results of merging RoBERTa models on 8 language tasks.}
\setlength{\tabcolsep}{2pt}
\begin{tabular}{l|cccccccc}
\toprule
\multirow{1}{*}{Methods} 
    & CoLA & SST2 & MRPC & STSB & QQP & MNLI & QNLI & RTE \\
\midrule
Individual & 60.2 & 94.0 & 89.2 & 90.6 & 91.4 & 87.2 & 92.7 & 79.1 \\
Pre-trained & 0.0 & 49.1 & 31.6 & 4.2 & 36.8 & 31.8 & 50.9 & 47.3 \\
\midrule
Weight Averaging  & 9.3 & 79.5 & 77.7 & 31.8 & 79.4 & 40.0 & 72.4 & 61.0 \\
Fisher Merging & 18.8 & 54.8 & 81.4 & 77.4 & 83.1 & 35.0 & 67.5 & 51.6 \\
RegMean & 36.7 & 90.6 & 75.7 & 62.7 & 83.6 & 70.0 & 82.4 & 58.5 \\
Task Arithmetic & 3.8 & 86.1 & 79.4 & 66.8 & 83.3 & 53.0 & 57.7 & 52.7 \\
Ties-Merging & 17.1 & 84.1 & 79.9 & 63.6 & 85.9 & 64.1 & 70.7 & 42.6 \\
Breadcrumbs & 34.3 & 86.2 & 80.1 & 60.3 & 83.3 & 58.8 & 79.2 & 63.5 \\
PCB-Merging & 23.5 & 84.6 & 79.4 & 68.9 & 82.0 & 63.6 & 76.6 & 64.5 \\
\midrule
Twin-Merging & 52.3 & 93.2 & 88.7 & 68.6 & 82.0 & 78.8 & 88.4 & 74.4 \\
EMR-Merging & 37.8 & 87.9 & 82.3 & 72.9 & 82.6 & 78.9 & 83.2 & 67.9 \\
\midrule
\textbf{MERGE-WA ($G_1$)} & 15.4$\pm$1.2 & 82.5$\pm$6.0 & 79.0$\pm$0.6 & 52.3$\pm$14.0 & 79.5$\pm$1.1 & 54.3$\pm$12.8 & 74.6$\pm$1.6 & 62.2$\pm$6.4 \\
\textbf{MERGE-WA ($G_2$)} & 57.0$\pm$3.3 & 90.7$\pm$3.0 & 81.6$\pm$2.3 & 73.9$\pm$2.5 & 86.8$\pm$3.1 & 83.0$\pm$3.5 & 86.3$\pm$2.9 & 72.9$\pm$2.2 \\
\textbf{MERGE-WA ($G_3$)} & \textbf{61.2$\pm$0.5} & \textbf{93.7$\pm$0.2} & \textbf{88.8$\pm$0.2} & \textbf{79.3$\pm$1.7} & \textbf{90.7$\pm$1.0} & \textbf{86.8$\pm$0.2} & \textbf{92.2$\pm$0.4} & \textbf{75.7$\pm$1.2} \\
\midrule
\textbf{MERGE-TA ($G_1$)} & 2.8$\pm$0.0 & 87.4$\pm$0.0 & 80.6$\pm$0.0 & 66.8$\pm$0.0 & 83.3$\pm$0.0 & 53.5$\pm$0.0 & 58.4$\pm$0.0 & 53.8$\pm$0.0 \\
\textbf{MERGE-TA ($G_2$)} & 50.4$\pm$5.5 & 90.5$\pm$1.7 & 82.4$\pm$1.2 & 77.3$\pm$3.2 & 88.7$\pm$2.2 & 81.4$\pm$4.8 & 89.7$\pm$1.9 & 69.2$\pm$3.2 \\
\textbf{MERGE-TA ($G_3$)} & \textbf{61.3$\pm$0.4} & \textbf{93.7$\pm$0.1} & \textbf{89.4$\pm$0.3} & \textbf{79.0$\pm$0.0} & \textbf{91.3$\pm$0.2} & \textbf{87.1$\pm$0.0} & \textbf{92.6$\pm$0.0} & \textbf{75.1$\pm$0.0} \\
\midrule
\textbf{MERGE-TM ($G_1$)} & 34.6$\pm$3.2 & 85.3$\pm$1.6 & 80.3$\pm$0.2 & 59.4$\pm$5.5 & 86.0$\pm$0.3 & 68.5$\pm$4.4 & 73.0$\pm$6.4 & 50.5$\pm$4.2 \\
\textbf{MERGE-TM ($G_2$)} & 51.2$\pm$5.9 & 91.2$\pm$2.6 & 83.7$\pm$2.0 & 76.6$\pm$3.5 & 88.8$\pm$1.3 & 83.5$\pm$2.1 & 88.9$\pm$2.8 & 71.3$\pm$2.0 \\
\textbf{MERGE-TM ($G_3$)} & \textbf{61.3$\pm$0.6} & \textbf{93.4$\pm$0.4} & 88.4$\pm$1.9 & \textbf{78.9$\pm$0.8} & \textbf{90.8$\pm$0.6} & \textbf{87.0$\pm$0.1} & \textbf{92.3$\pm$0.4} & \textbf{76.1$\pm$1.3} \\
\midrule
\textbf{MERGE-BC ($G_1$)} & 32.8$\pm$1.4 & 84.6$\pm$0.8 & 79.6$\pm$0.1 & 63.7$\pm$6.9 & 83.8$\pm$0.1 & 80.4$\pm$8.7 & 78.9$\pm$0.2 & 64.0$\pm$0.3 \\
\textbf{MERGE-BC ($G_2$)} & 60.0$\pm$0.2 & 91.3$\pm$2.7 & 81.8$\pm$3.6 & 75.0$\pm$2.2 & 87.1$\pm$2.6 & 80.1$\pm$6.1 & 88.2$\pm$4.1 & 69.0$\pm$2.8 \\
\textbf{MERGE-BC ($G_3$)} & \textbf{61.3$\pm$0.4} & \textbf{93.8$\pm$0.1} & \textbf{89.4$\pm$0.2} & \textbf{79.0$\pm$0.0} & \textbf{91.4$\pm$0.0} & \textbf{87.1$\pm$0.0} & \textbf{92.5$\pm$0.1} & \textbf{75.0$\pm$0.1} \\
\bottomrule
\end{tabular}
\label{tab:RoBERTa}
\end{table}

\subsubsection{Merging Vision Models}

As shown in Table~\ref{tab:main_results}, Table~\ref{tab:ViTB32} and Table~\ref{tab:ViTL14}, for both ViT-B/32 and ViT-L/14 vision models, our storage-minimal solutions ($G_1$) outperform their respective base methods while requiring only about 30\% of the storage. Notably, our cost-effective solutions ($G_2$) exceed the performance of the best merging baseline, EMR-Merging, with comparable or superior storage efficiency. At the high-performance end, our $G_3$ solutions nearly match the accuracy of individual models at significantly lower cost, a level unattainable by any other merging technique. Fig.~\ref{fig:all}\subref{fig3:sub1} and \ref{fig:all}\subref{fig3:sub2} visualize the comparative results, where MERGE-TM produces superior fronts, demonstrating its consistent effectiveness across different vision model scales.

\begin{table}[htbp]
\tiny
\centering
\caption{Per-task results of merging GPT-2 models on 7 language tasks.}
\setlength{\tabcolsep}{3pt}
\begin{tabular}{l|ccccccc}
\toprule
\multirow{1}{*}{Methods} 
    & CoLA & SST2 & MRPC & QQP & MNLI & QNLI & RTE \\
\midrule
Individual & 76.8 & 91.2 & 80.4 & 89.6 & 82.1 & 88.3 & 65.3 \\
Pre-trained & 30.9 & 50.9 & 31.4 & 63.2 & 33.0 & 49.2 & 52.7 \\
\midrule
Weight Averaging & 55.0 & 52.5 & 51.0 & 76.7 & 55.1 & 57.6 & 44.8 \\
Fisher Merging & 54.8 & 64.7 & 39.5 & 81.5 & 58.0 & 63.3 & 49.1 \\
RegMean & 61.7 & 79.7 & 65.4 & 78.8 & 70.4 & 69.7 & 56.0 \\
Task Arithmetic & 69.0 & 82.0 & 70.1 & 81.7 & 67.4 & 70.2 & 47.3 \\
Ties-Merging & 63.5 & 71.1 & 43.4 & 83.4 & 72.9 & 70.9 & 48.7 \\
Breadcrumbs & 67.5 & 76.3 & 69.9 & 80.9 & 66.5 & 65.5 & 45.8 \\
PCB-Merging & 70.0 & 82.2 & 72.1 & 78.1 & 66.9 & 68.4 & 50.5 \\
\midrule
Twin-Merging & 74.4 & 88.8 & 76.7 & 89.5 & 81.6 & \textbf{88.7} & 64.3 \\
EMR-Merging & 72.8 & \textbf{90.3} & 79.2 & 88.1 & 81.1 & 84.8 & \textbf{66.5} \\
\midrule
\textbf{MERGE-WA ($G_1$)} & 55.7$\pm$2.1 & 79.9$\pm$11.5 & 60.7$\pm$6.8 & 77.6$\pm$0.5 & 59.6$\pm$4.5 & 64.6$\pm$6.6 & 46.4$\pm$1.4 \\
\textbf{MERGE-WA ($G_2$)} & 75.4$\pm$0.7 & 88.8$\pm$0.5 & 80.0$\pm$0.2 & 89.0$\pm$0.6 & 81.9$\pm$0.3 & 88.2$\pm$0.1 & 63.8$\pm$1.8 \\
\textbf{MERGE-WA ($G_3$)} & \textbf{76.0$\pm$1.0} & 88.8$\pm$0.6 & \textbf{80.1$\pm$0.3} & 89.2$\pm$0.7 & \textbf{81.9$\pm$0.3} & 88.3$\pm$0.1 & 64.5$\pm$1.9 \\
\midrule
\textbf{MERGE-TA ($G_1$)} & 68.9$\pm$0.1 & 83.5$\pm$0.0 & 70.6$\pm$0.4 & 81.6$\pm$0.0 & 67.4$\pm$0.0 & 70.2$\pm$0.0 & 47.3$\pm$0.0 \\
\textbf{MERGE-TA ($G_2$)} & 75.2$\pm$1.4 & 88.1$\pm$1.0 & 78.9$\pm$1.2 & 88.7$\pm$0.7 & 81.7$\pm$0.4 & 88.2$\pm$0.2 & 64.8$\pm$1.5 \\
\textbf{MERGE-TA ($G_3$)} & \textbf{76.9$\pm$0.1} & 88.9$\pm$0.3 & \textbf{80.1$\pm$0.4} & 89.4$\pm$0.1 & \textbf{82.1$\pm$0.1} & 88.2$\pm$0.2 & 65.3$\pm$0.0 \\
\midrule
\textbf{MERGE-TM ($G_1$)} & 62.1$\pm$1.8 & 84.1$\pm$5.6 & 41.6$\pm$2.5 & 84.5$\pm$0.5 & 74.4$\pm$0.4 & 74.7$\pm$6.9 & 50.4$\pm$1.6 \\
\textbf{MERGE-TM ($G_2$)} & 75.5$\pm$1.5 & 88.0$\pm$1.4 & 80.4$\pm$1.2 & 88.7$\pm$0.8 & 81.4$\pm$0.4 & 87.6$\pm$1.3 & 64.7$\pm$2.0 \\
\textbf{MERGE-TM ($G_3$)} & \textbf{76.7$\pm$0.2} & 88.8$\pm$0.4 & \textbf{80.6$\pm$0.9} & \textbf{89.5$\pm$0.0} & \textbf{82.1$\pm$0.0} & 88.3$\pm$0.1 & 65.3$\pm$0.2 \\
\midrule
\textbf{MERGE-BC ($G_1$)} & 66.5$\pm$0.3 & 77.2$\pm$0.0 & 71.4$\pm$0.1 & 81.6$\pm$0.0 & 65.9$\pm$0.0 & 87.0$\pm$0.0 & 48.4$\pm$0.0 \\
\textbf{MERGE-BC ($G_2$)} & 76.1$\pm$0.4 & 88.2$\pm$0.6 & 78.8$\pm$1.0 & 86.6$\pm$1.0 & 81.7$\pm$0.7 & 87.1$\pm$1.6 & 67.1$\pm$1.2 \\
\textbf{MERGE-BC ($G_3$)} & \textbf{76.5$\pm$0.2} & 88.2$\pm$0.5 & \textbf{80.0$\pm$0.9} & \textbf{89.5$\pm$0.1} & \textbf{82.1$\pm$0.0} & 88.2$\pm$0.1 & 65.8$\pm$0.5 \\
\bottomrule
\end{tabular}
\label{tab:GPT-2}
\end{table}

\subsubsection{Merging FFT Language Models}

MERGE again demonstrates significant advantages over the baselines in language tasks. Results in Table~\ref{tab:main_results}, Table~\ref{tab:RoBERTa} and Table~\ref{tab:GPT-2} reveal that our method consistently yields superior solutions compared to the top-performing dynamic baseline, Twin-Merging. For instance, on RoBERTa, all $G_2$ solutions achieve both performance gains and cost reductions. Furthermore, our $G_3$ solutions surpass all merging baselines by over 5.2\% for RoBERTa and 0.7\% for GPT-2. Crucially, these solutions substantially narrow the performance gap to task-specific models, reducing the drop from over 7.3\% to below 2.1\% for RoBERTa, and from over 1.4\% to below 0.7\% for GPT-2. Fig.~\ref{fig:all}\subref{fig3:sub3} and \ref{fig:all}\subref{fig3:sub4} highlight the capabilities of Task Arithmetic and Ties-Merging equipped with MERGE, validating the extensibility of MERGE to fully fine-tuned language models.

\begin{table*}[t]
% \tiny
\scriptsize
\centering
\caption{Per-task results of merging PEFT models on 11 language tasks.}
\setlength{\tabcolsep}{3.0pt}
\begin{tabular}{l|ccccccccccc}
\toprule
\multirow{1}{*}{Methods} 
    & RTE & CB & Winogrande & WiC & WSC & COPA & H-SWAG & Story Cloze & ANLI-R1 & ANLI-R2 & ANLI-R3 \\
\midrule
Individual & 82.7 & 95.8 & 75.1 & 71.7 & 65.3 & 85.3 & 44.4 & 94.9 & 70.2 & 46.5 & 53.0 \\
Traditional MTL & 88.6 & 95.8 & 75.5 & 61.1 & 80.6 & 94.1 & 42.3 & 97.6 & 70.5 & 49.8 & 47.7 \\
Pre-trained & 58.4 & 54.2 & 51.2 & 51.9 & 63.9 & 75 & 39 & 86.5 & 35.5 & 34.4 & 34.3 \\
\midrule
Weight Averaging  & 81.2 & 58.3 & 53.8 & 55.2 & 53.5 & 80.9 & 40.1 & 92.5 & 43.3 & 39.2 & 40.2 \\
Fisher Merging & 83.3 & 83.3 & 56.7 & 54.2 & 58.3 & 83.1 & 42.2 & 94.1 & 45.9 & 41.0 & 42.2 \\
RegMean & 81.2 & 58.3 & 53.8 & 55.2 & 53.5 & 80.9 & 40.1 & 92.5 & 43.3 & 39.2 & 40.2 \\
Task Arithmetic & 76.5 & 79.2 & 57.7 & 51.6 & 51.4 & 66.2 & 31.4 & 81.5 & 59.8 & 47.5 & 48.2 \\
Ties-Merging & 81.2 & 87.5 & 60.8 & 59.9 & 58.3 & 80.2 & 42.6 & 91.1 & 58.1 & 46.5 & 47.4 \\
Breadcrumbs & 76.7 & 62.5 & 57.8 & 54.6 & 46.5 & 83.1 & 41.3 & 93.4 & 45.2 & 39.9 & 41.4 \\
PCB-Merging & \textbf{85.9} & 83.3 & 61.9 & 57.1 & 63.9 & 82.4 & 42.7 & 91.2 & 64.2 & 47.8 & 45.9 \\
\midrule
Twin-Merging & 81.2 & 85.6 & 65.4 & 57.2 & \textbf{66.3} & 81.6 & 44.4 & 92.9 & 66.5 & 42.4 & \textbf{51.2} \\
EMR-Merging & 81.8 & 87.5 & 66.6 & 56.1 & 65.3 & 82.4 & \textbf{44.7} & 93.6 & 65.7 & 43.8 & 50.8 \\
\midrule
\textbf{MERGE-WA ($G_1$)} & 75.9$\pm$4.3 & 82.5$\pm$8.1 & 52.6$\pm$0.9 & 51.7$\pm$0.1 & 65.0$\pm$0.3 & 79.1$\pm$0.9 & 39.4$\pm$0.0 & 91.2$\pm$0.4 & 42.2$\pm$1.4 & 38.6$\pm$0.8 & 37.2$\pm$1.3 \\
\textbf{MERGE-WA ($G_2$)} & 82.1$\pm$4.5 & 91.7$\pm$0.0 & 69.8$\pm$4.0 & 69.7$\pm$2.2 & 64.2$\pm$0.6 & 81.7$\pm$1.0 & 41.6$\pm$0.9 & 92.3$\pm$0.4 & 60.5$\pm$2.8 & 45.5$\pm$0.9 & 45.2$\pm$1.0 \\
\textbf{MERGE-WA ($G_3$)} & 78.0$\pm$1.9 & \textbf{95.8$\pm$0.0} & \textbf{73.9$\pm$0.7} & \textbf{72.5$\pm$0.8} & 63.9$\pm$0.0 & 81.1$\pm$0.3 & 43.8$\pm$0.3 & \textbf{94.1$\pm$0.6} & 65.9$\pm$1.7 & \textbf{46.6$\pm$0.8} & 48.4$\pm$1.2 \\
\midrule
\textbf{MERGE-TA ($G_1$)} & 77.9$\pm$0.3 & 75.0$\pm$0.0 & 59.7$\pm$0.4 & 49.7$\pm$0.1 & 56.6$\pm$0.3 & 84.5$\pm$0.8 & 40.3$\pm$0.4 & 92.9$\pm$0.3 & 62.6$\pm$0.3 & 48.7$\pm$0.1 & 49.6$\pm$0.1 \\
\textbf{MERGE-TA ($G_2$)} & 83.1$\pm$3.4 & 88.3$\pm$3.1 & 68.1$\pm$3.5 & 63.0$\pm$5.6 & 61.8$\pm$4.8 & 85.6$\pm$1.2 & 42.0$\pm$0.6 & 95.5$\pm$0.6 & 66.8$\pm$0.9 & 49.3$\pm$0.4 & 51.0$\pm$1.1 \\
\textbf{MERGE-TA ($G_3$)} & 84.4$\pm$3.3 & \textbf{89.2$\pm$3.4} & \textbf{68.8$\pm$3.4} & \textbf{63.1$\pm$5.7} & 63.9$\pm$2.3 & \textbf{85.4$\pm$1.2} & 42.4$\pm$0.7 & \textbf{95.6$\pm$0.6} & \textbf{67.0$\pm$1.1} & \textbf{49.0$\pm$0.6} & 51.1$\pm$1.0 \\
\midrule
\textbf{MERGE-TM ($G_1$)} & 84.4$\pm$0.3 & 86.7$\pm$3.1 & 64.5$\pm$3.5 & 57.5$\pm$5.4 & 63.1$\pm$1.1 & 78.1$\pm$0.7 & 42.7$\pm$0.2 & 90.1$\pm$0.7 & 62.0$\pm$0.6 & 45.9$\pm$0.2 & 45.8$\pm$0.1 \\
\textbf{MERGE-TM ($G_2$)} & 82.3$\pm$1.7 & 92.5$\pm$1.6 & 68.1$\pm$2.6 & 65.6$\pm$5.2 & 64.7$\pm$1.1 & 79.4$\pm$1.8 & 43.2$\pm$0.1 & 91.8$\pm$1.4 & 64.7$\pm$1.4 & 46.3$\pm$0.8 & 47.3$\pm$1.0 \\
\textbf{MERGE-TM ($G_3$)} & 81.2$\pm$1.1 & \textbf{95.0$\pm$1.6} & \textbf{72.7$\pm$1.3} & \textbf{70.9$\pm$1.1} & 63.9$\pm$0.0 & 81.0$\pm$1.7 & 43.7$\pm$0.3 & 93.1$\pm$0.9 & 65.8$\pm$1.7 & \textbf{45.6$\pm$1.6} & 50.3$\pm$1.1 \\
\midrule
\textbf{MERGE-BC ($G_1$)} & 84.5$\pm$0.3 & 87.5$\pm$0.0 & 57.4$\pm$3.5 & 53.8$\pm$0.4 & 61.2$\pm$2.3 & 81.2$\pm$0.4 & 40.7$\pm$0.0 & 92.4$\pm$0.1 & 50.4$\pm$0.4 & 43.1$\pm$0.2 & 43.1$\pm$0.2 \\
\textbf{MERGE-BC ($G_2$)} & 82.4$\pm$3.1 & 93.3$\pm$2.0 & 73.2$\pm$2.5 & 68.8$\pm$3.5 & 63.9$\pm$0.0 & 78.4$\pm$1.8 & 43.4$\pm$0.8 & 93.7$\pm$0.4 & 62.2$\pm$2.9 & 43.3$\pm$2.0 & 46.1$\pm$2.9 \\
\textbf{MERGE-BC ($G_3$)} & 74.4$\pm$4.2 & \textbf{95.8$\pm$0.0} & \textbf{74.5$\pm$0.1} & \textbf{72.4$\pm$0.9} & 63.9$\pm$0.0 & 80.9$\pm$0.5 & 44.0$\pm$0.3 & \textbf{94.2$\pm$0.3} & \textbf{67.2$\pm$2.2} & \textbf{45.3$\pm$0.7} & 49.4$\pm$1.2 \\
\bottomrule
\end{tabular}
\label{tab:PEFT}
\end{table*}

\subsubsection{Merging PEFT Models}

As shown in Table~\ref{tab:main_results} and Table~\ref{tab:PEFT}, the solutions generated by MERGE variants are highly storage-efficient. In terms of performance, they consistently outperform their base methods and, in many cases, also surpass the best merging baseline. As illustrated in Fig.~\ref{fig:all}\subref{fig3:sub5}, MERGE-TA is particularly notable for its Pareto front efficiency, while other variants prove more adept at discovering a broader set of high-performance solutions. These findings confirm the efficacy of MERGE for PEFT models.

\subsection{MERGE Analysis}
\label{subsec:MERGE Analysis}
\subsubsection{Unseen Generalization}

To evaluate the generalization ability of MERGE in transferring learned knowledge to novel tasks, we compare its performance against several merging baselines on a set of unseen tasks. Specifically, we merge five out of eight expert vision models and evaluate the resulting solutions on the corresponding five source tasks as well as three unseen tasks (i.e., GTSRB, MNIST, and DTD). As summarized in Table~\ref{tab:unseen}, the top-performing representative solution not only achieves significantly lower storage requirements but also consistently outperforms other merging methods by 1.3\%–4.2\% on the three unseen vision tasks and by 2.5\%–9.3\% on the five source tasks. Meanwhile, the most storage-efficient representative solution remains competitive with Ties-Merging while requiring only about one quarter of the storage. These results demonstrate that MERGE effectively enhances generalization to novel tasks without compromising performance on the source tasks.

\begin{table}[htbp]
\scriptsize
\centering
\caption{Results on five seen and three unseen tasks.}
\setlength{\tabcolsep}{7.0pt}
\begin{tabular}{l|c|cc}
\toprule
\multirow{2}{*}{Methods}
    & \multirow{2}{*}{$\mathcal{C}$ ($\downarrow$)} 
    & \multicolumn{2}{c}{$\mathcal{P}$ ($\uparrow$)} \\
\cmidrule(lr){3-4}
    & & Seen Tasks & Unseen Tasks \\
\midrule
Weight Averaging & 1.00 & 72.8 & 50.3 \\
Task Arithmetic & 1.00 & 77.0 & 50.8 \\
Task Arithmetic w/ DARE & 1.00 & 73.9 & 50.7 \\
Breadcrumbs & 1.00 & 76.8 & 53.0 \\
Ties-Merging & 1.00 & \underline{79.6} & 52.5 \\
Ties-Merging w/ DARE & 1.00 & 79.2 & \underline{53.2} \\
\midrule
\textbf{MERGE-TM (min($\mathcal{C}$))} & \textbf{0.2513} & \underline{79.6} & 52.5 \\
\textbf{MERGE-TM (max($\mathcal{P}$))} & \underline{0.3141} & \textbf{82.1} & \textbf{54.5} \\
\bottomrule
\end{tabular}
\label{tab:unseen}
\end{table}

\begin{figure*}[htbp]
\centering
\subfloat[]{%
    \includegraphics[width=0.33\linewidth]{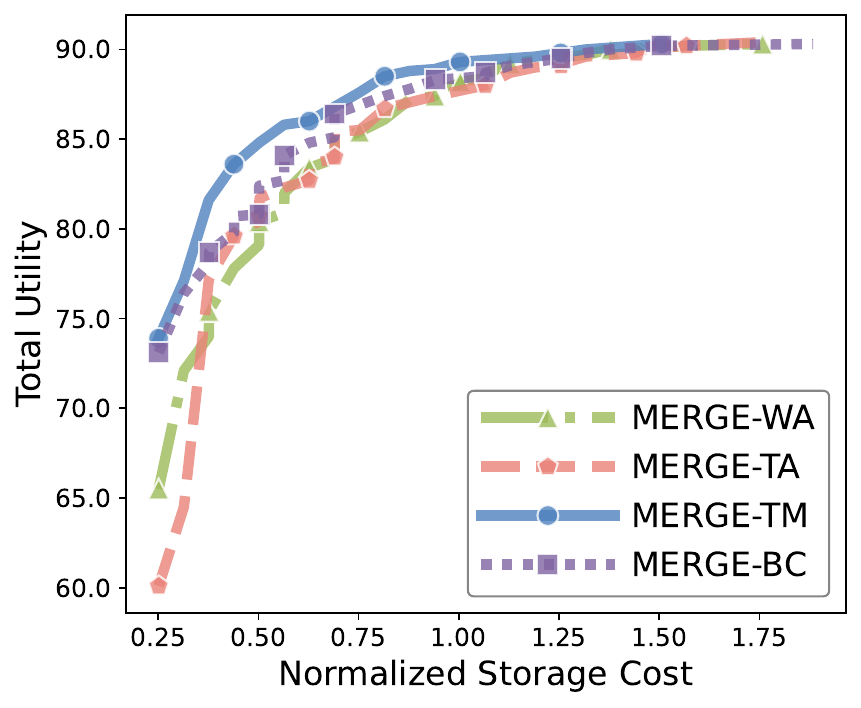}%
    \label{fig4:sub1}}
\hfil
\subfloat[]{%
    \includegraphics[width=0.33\linewidth]{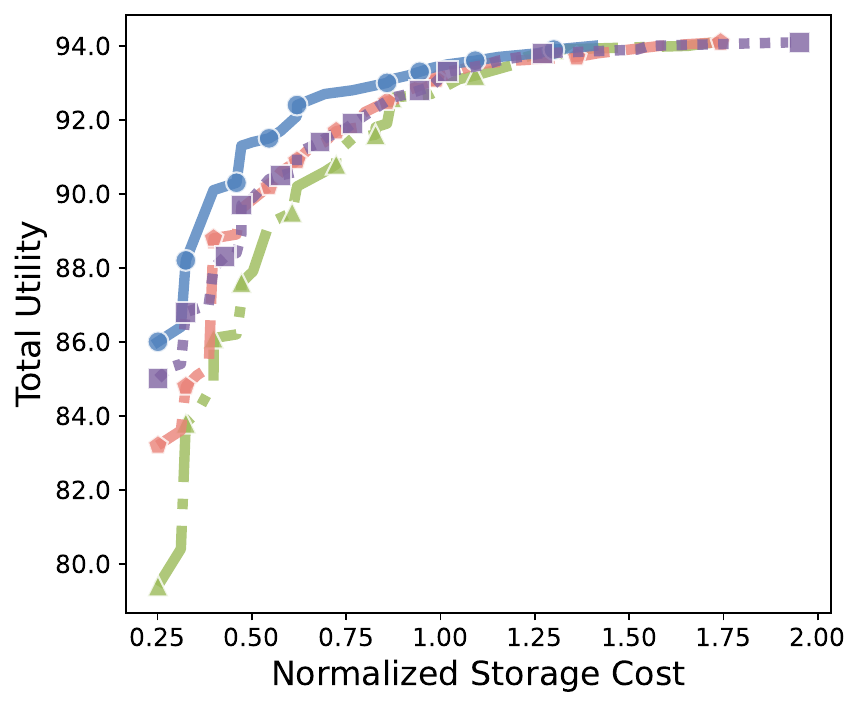}%
    \label{fig4:sub2}}
\hfil
\subfloat[]{%
    \includegraphics[width=0.33\linewidth]{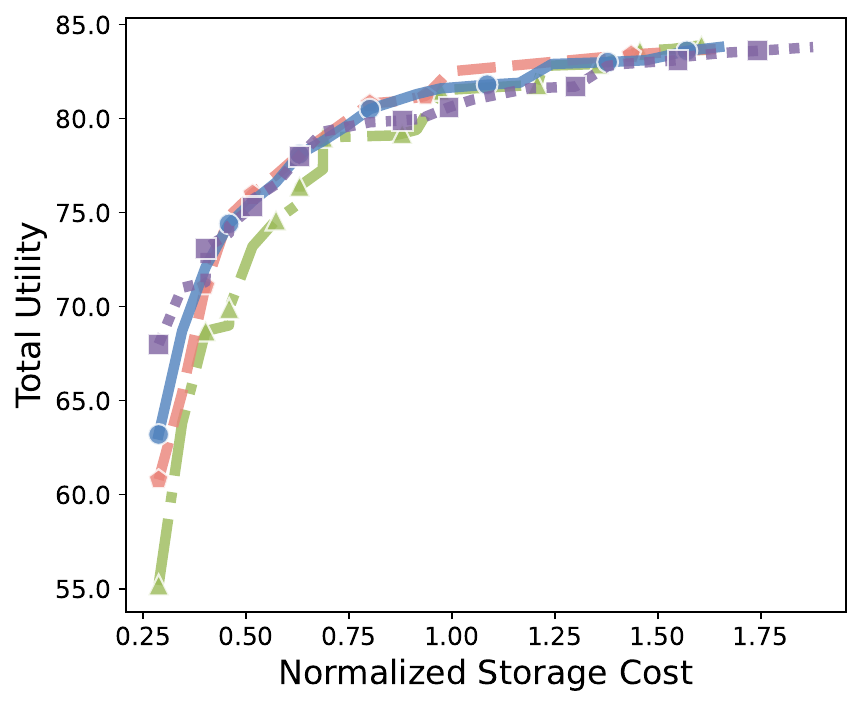}%
    \label{fig4:sub3}}
\vfill
\subfloat[]{%
    \includegraphics[width=0.33\linewidth]{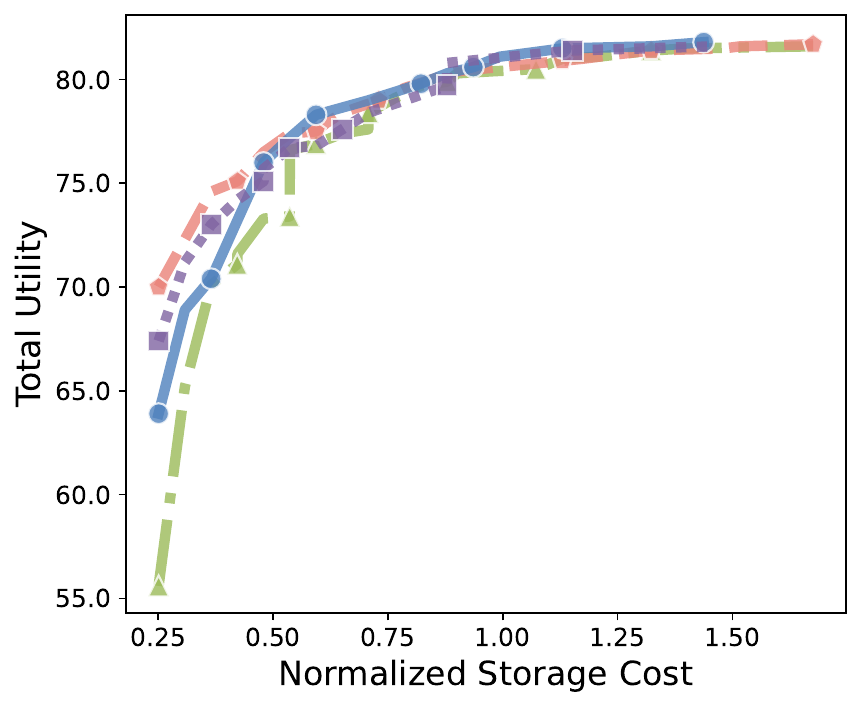}%
    \label{fig4:sub4}}
\hfil
\subfloat[]{%
    \includegraphics[width=0.33\linewidth]{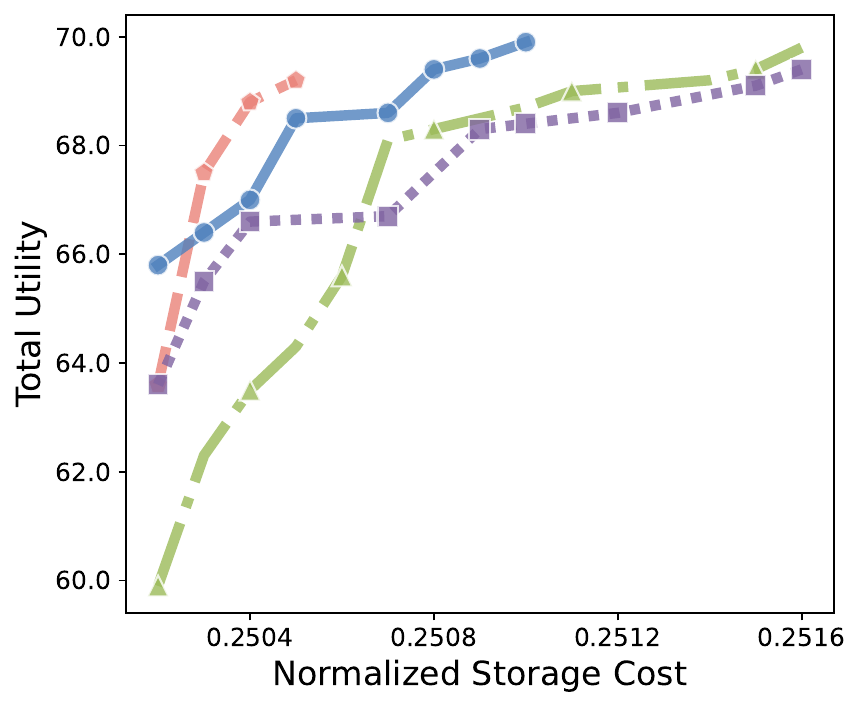}%
    \label{fig4:sub5}}
\caption{Merging utility of MERGE on various models. (a) ViT-B/32 models. (b) ViT-L/14 models. (c) RoBERTa models. (d) GPT-2 models. (e) PEFT models.}
\label{fig:merging_utility}
\end{figure*}

\subsubsection{Merging Utility}

This analysis investigates the relationship between storage cost and performance gains across different models and tasks. Our findings reveal a clear pattern of diminishing returns in fine-grained merging. As shown in Fig.~\ref{fig:merging_utility}, the total utility curves are initially steep in the low-cost region but gradually flatten as the storage resource increases. This indicates that the marginal utility decreases with higher storage cost. To statistically validate this negative correlation, we use Spearman's rank correlation coefficient \cite{hollander2013nonparametric}. Among 20 cases analyzed, 18 demonstrate statistical significance ($p<0.05$), with detailed results provided in Table~\ref{tab: merging utility}. Notably, this correlation is generally more pronounced for language models, such as RoBERTa and GPT-2, than for vision models, indicating that performance gains from increased storage tend to saturate more quickly when merging fully fine-tuned language models.

\begin{table}[htbp]
\scriptsize
\centering
\caption{Spearman rank-order correlations between normalized storage cost and marginal utility. The number of symbol '*' indicates significance levels: '*' for $p<0.05$, '**' for $p<0.01$, and '***' for $p<0.001$.}
\setlength{\tabcolsep}{2.5pt}
\begin{tabular}{l|c|c|c|c|c}
\toprule
Methods & ViT-B/32 & ViT-L/14 & RoBERTa & GPT-2 & PEFT\\
\midrule
MERGE-WA & $-0.69^{***}$ & $-0.58^{***}$ & $-0.60^{**}\phantom{*}$ & $-0.72^{***}$ & $-0.68^{*}\phantom{**}$ \\
MERGE-TA & $-0.62^{***}$ & $-0.61^{***}$ & $-0.81^{***}$ & $-0.79^{***}$ & $-1.00^{***}$ \\
MERGE-TM & $-0.60^{**}\phantom{*}$  & $-0.71^{***}$ & $-0.86^{***}$ & $-0.92^{***}$ & $-0.38^{\phantom{***}}$ \\
MERGE-BC      & $-0.63^{***}$ & $-0.58^{***}$ & $-0.67^{***}$ & $-0.57^{*}\phantom{**}$   & $-0.42^{\phantom{***}}$ \\
\bottomrule
\end{tabular}
\label{tab: merging utility}
\end{table}

\subsubsection{Component-Association Networks}

\begin{figure*}[htbp]
\centering
\subfloat[]{%
    \includegraphics[width=0.48\linewidth]{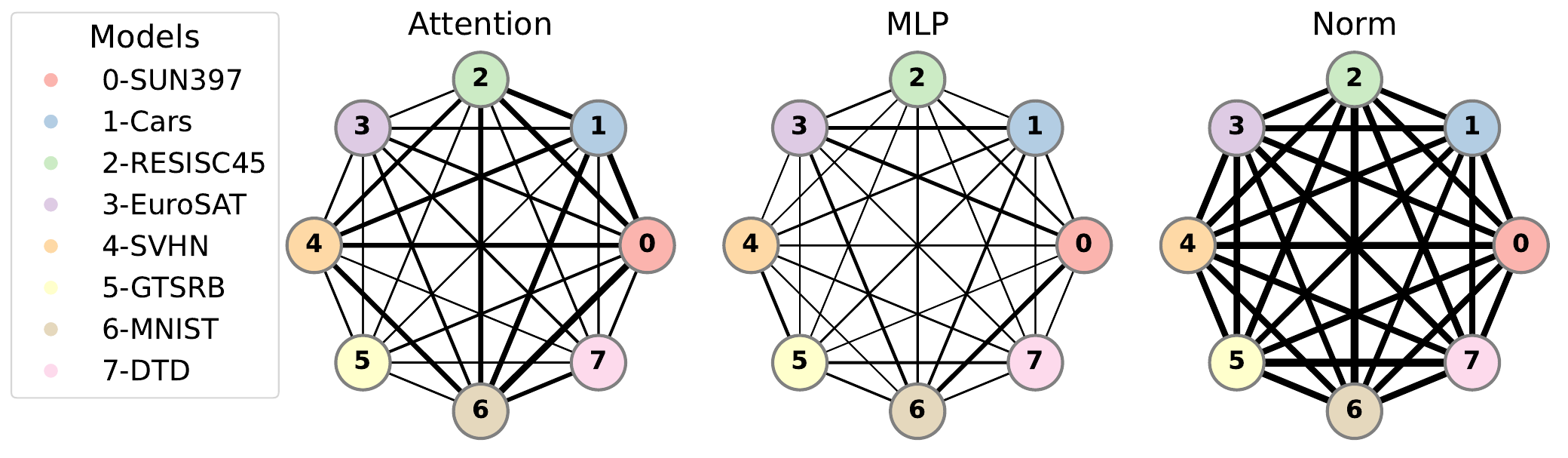}%
    \label{fig5:sub1}}
\hfil
\subfloat[]{%
    \includegraphics[width=0.48\linewidth]{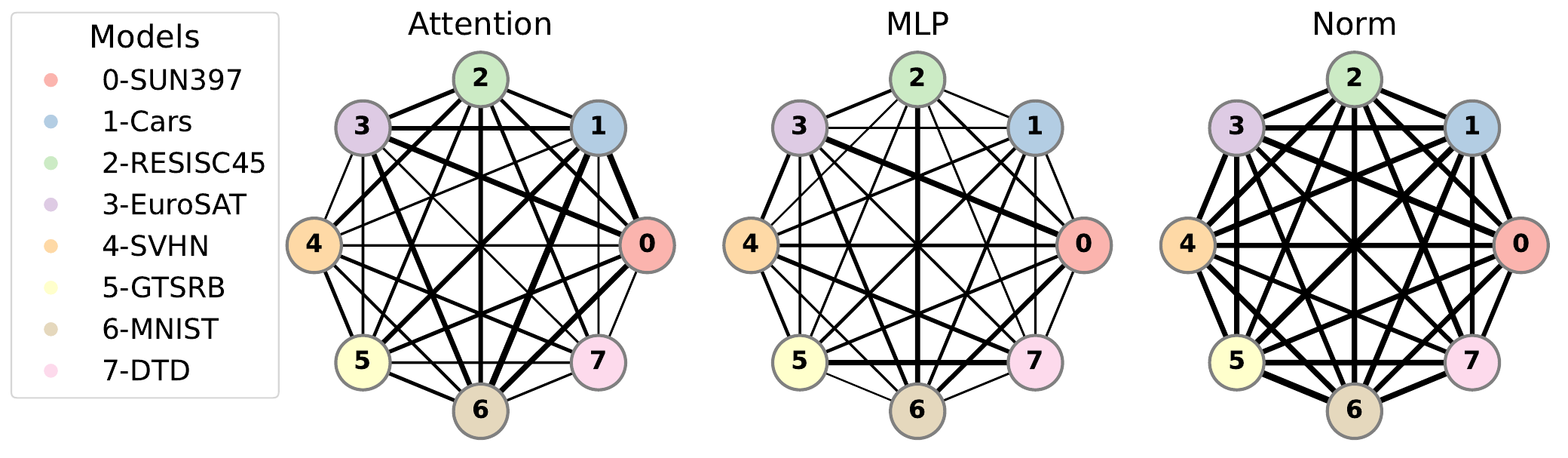}%
    \label{fig5:sub2}}
\vfill
\subfloat[]{%
    \includegraphics[width=0.48\linewidth]{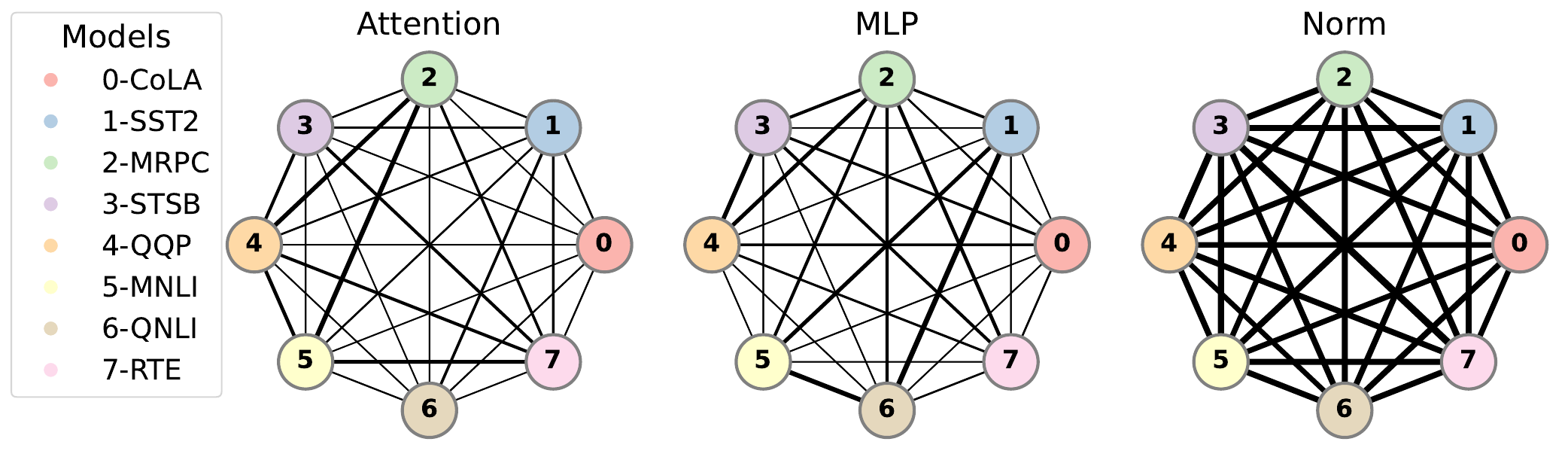}%
    \label{fig5:sub3}}
\hfil
\subfloat[]{%
    \includegraphics[width=0.48\linewidth]{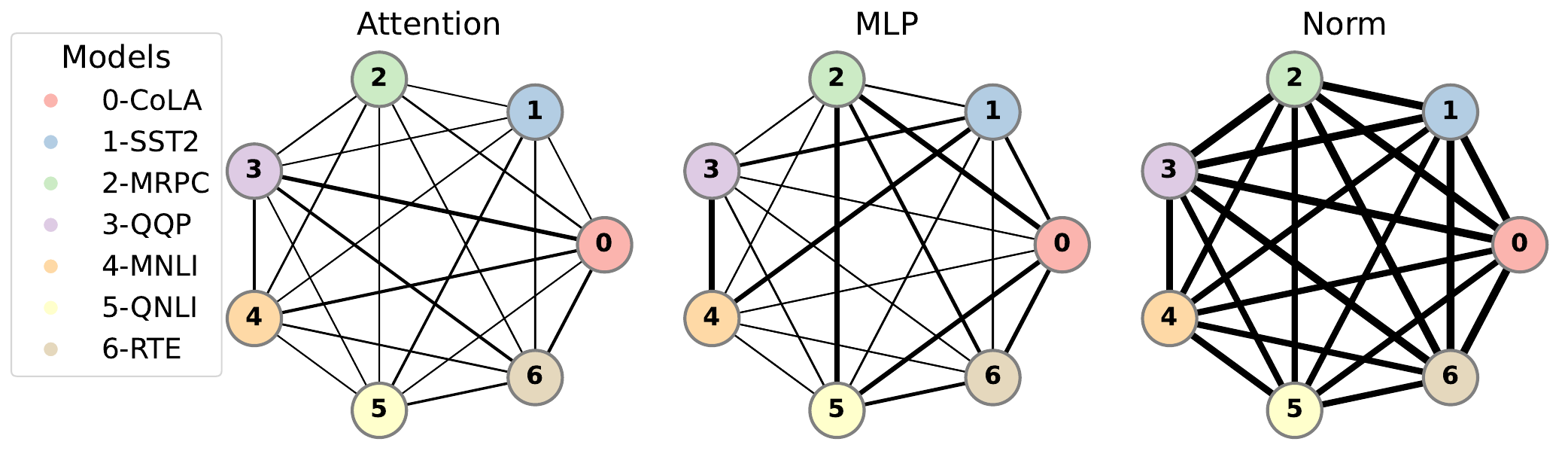}%
    \label{fig5:sub4}}
\vfill
\subfloat[]{%
    \includegraphics[width=0.7\linewidth]{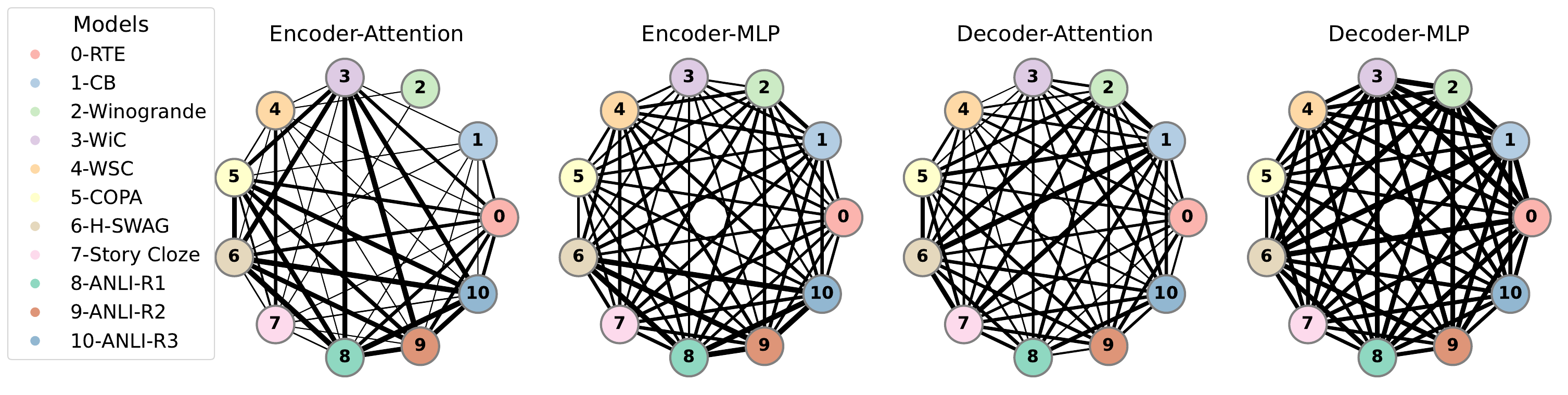}%
    \label{fig5:sub5}}
\caption{Relation networks of various models by components. (a) ViT-B/32 models on 8 tasks. (b) ViT-L/14 models on 8 tasks. (c) RoBERTa models on 8 tasks. (d) GPT-2 models on 7 tasks. (e) PEFT models on 11 tasks.}
\label{fig:relation_networks}
\end{figure*}

To investigate the intrinsic mechanisms of fine-grained merging, we construct component-association networks in which nodes represent homologous components of task-specific models and edge weights reflect their co-occurrence frequency across all discovered merging configurations. As illustrated in Fig.~\ref{fig:relation_networks}, these networks reveal key patterns of knowledge integration. First, normalization layers consistently exhibit universal knowledge sharing across both vision and language tasks, underscoring their robustness in merging. Second, integration preferences vary across domains. Specifically, ViT models predominantly rely on MLP layers for specialization, whereas RoBERTa and GPT-2 exhibit a notable preference for attention layers. Finally, functional specialization is position-dependent within the encoder-decoder architecture. For instance, in T0-3B PEFT models, Decoder-MLP modules actively contribute to knowledge sharing, while Encoder-Attention and Encoder-MLP modules tend to remain task-specific. These findings offer novel insights into the dynamics of fine-grained model merging, shedding light on how task-specific components interact and integrate knowledge across domains.

\subsection{Ablation Study}
\label{subsec:Ablation Study}
\subsubsection{Merging Granularity}

\begin{figure}[htbp]
\centering
\subfloat[]{%
    \includegraphics[width=0.48\linewidth]{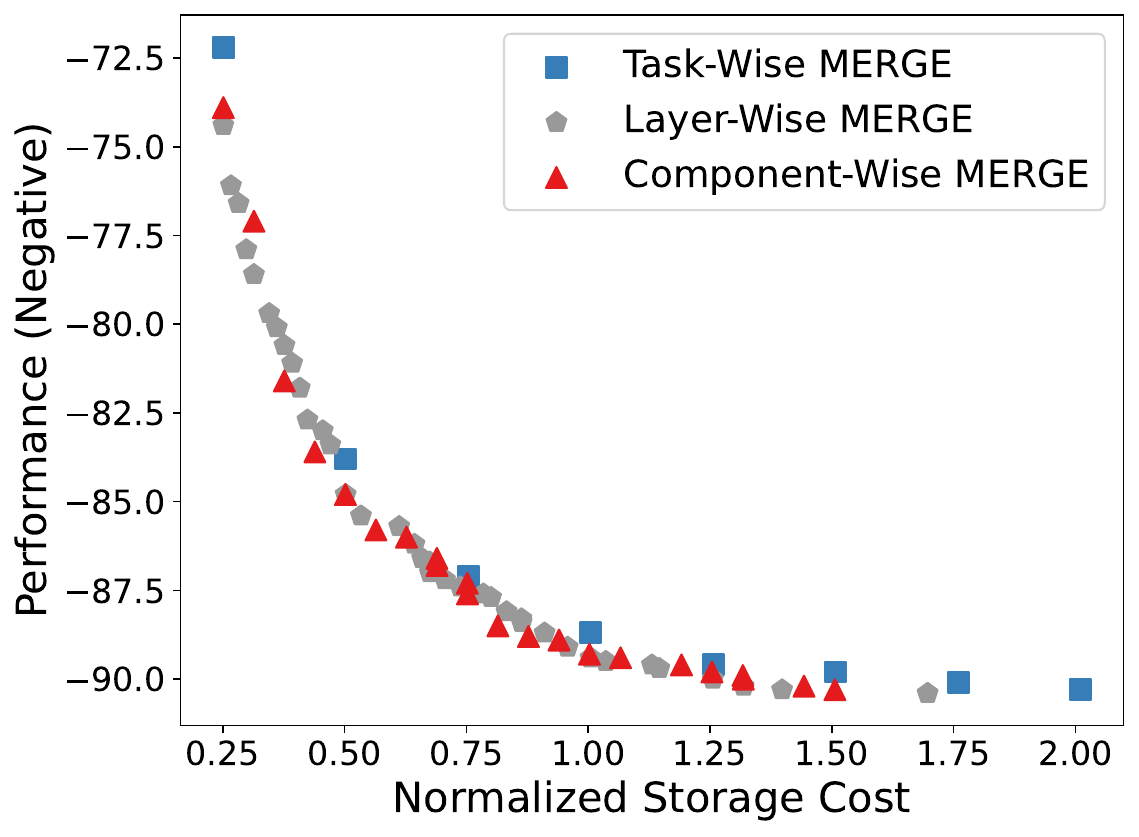}%
    \label{fig6:sub1}}
\hfil
\subfloat[]{%
    \includegraphics[width=0.48\linewidth]{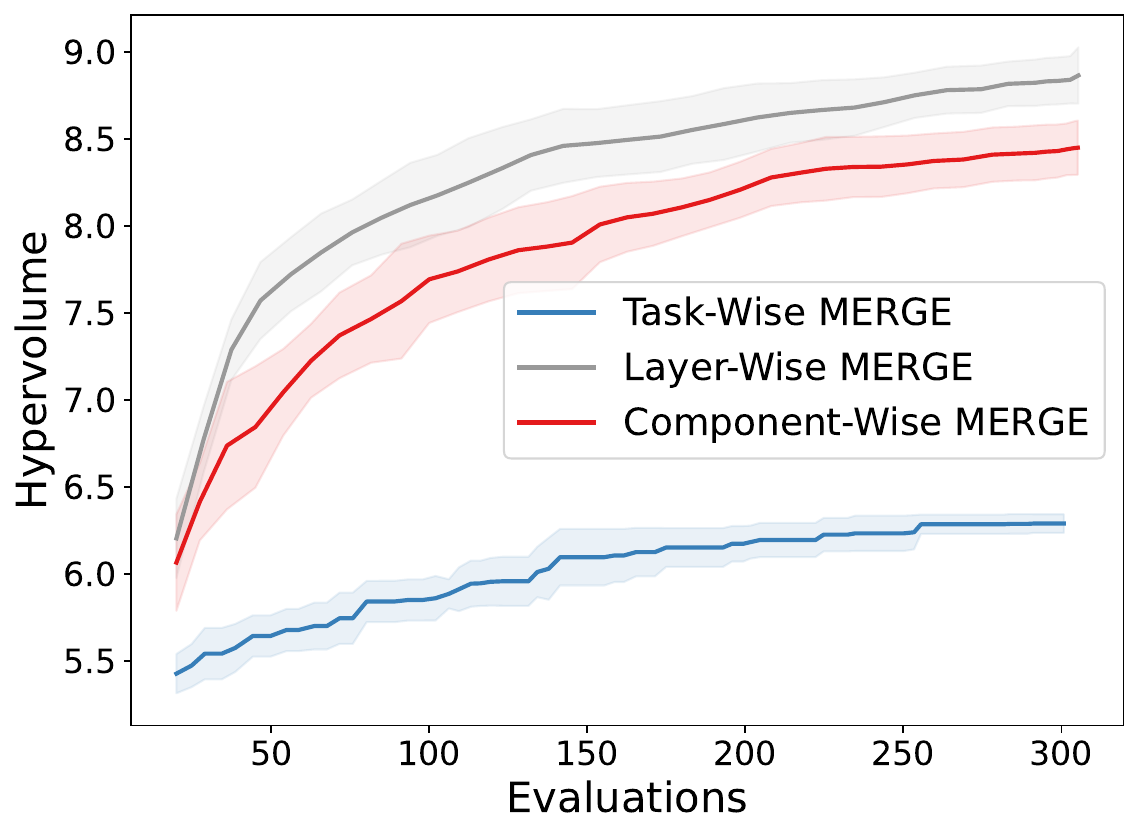}%
    \label{fig6:sub2}}
\caption{Merging granularity comparison. (a) Granularity: solutions. (b) Hypervolume evolution.}
\label{fig:granularity}
\end{figure}

Our component-wise merging method performs better than task-wise and layer-wise alternatives. As shown in Fig.~\ref{fig:granularity}\subref{fig6:sub1}, our method provides higher-quality solutions that achieve superior performance at lower cost. In contrast, the task-wise approach suffers from premature convergence during optimization (Fig.~\ref{fig:granularity}\subref{fig6:sub2}), while the layer-wise approach yields inferior solutions. This study confirms the effectiveness of our component-wise fine-grained merging.

\subsubsection{Surrogate-Assisted Optimization}
\begin{figure}[htbp]
\centering
\subfloat[]{%
    \includegraphics[width=0.48\linewidth]{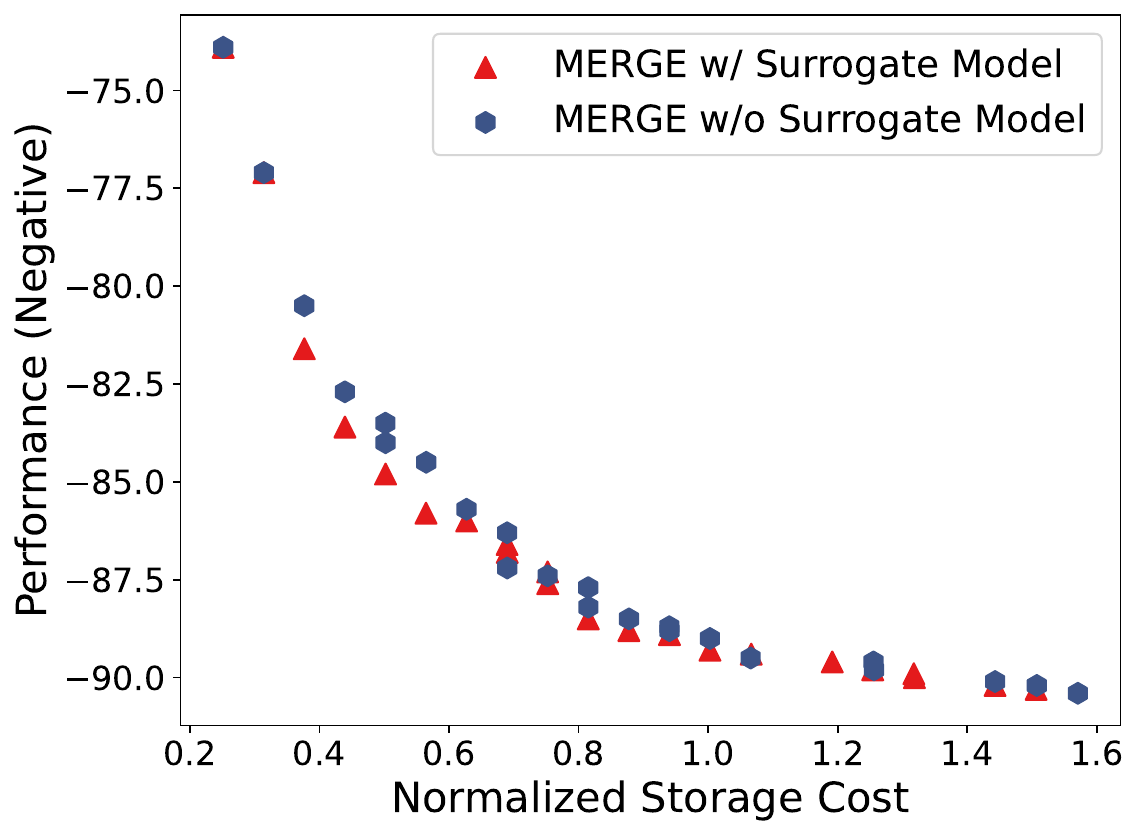}%
    \label{fig7:sub1}}
\hfil
\subfloat[]{%
    \includegraphics[width=0.48\linewidth]{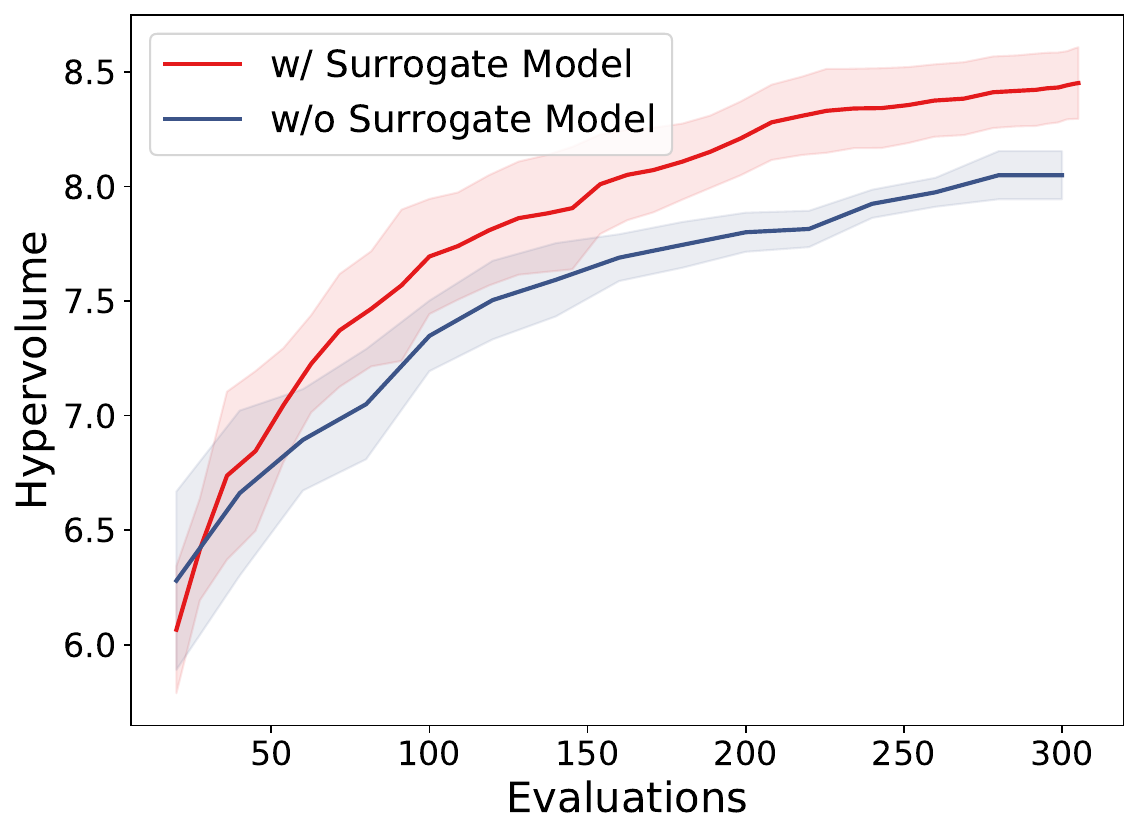}%
    \label{fig7:sub2}}
\caption{The impact of the surrogate model in MERGE. (a) Pareto-optimal solutions. (b) Hypervolume evolution.}
\label{fig:surrogate_model}
\end{figure}

To validate the efficacy of surrogate-assisted optimization, we compare MERGE against the version without the surrogate model. The surrogate-guided search discovers a significantly higher-quality Pareto front (Fig.~\ref{fig:surrogate_model}\subref{fig7:sub1}), despite generating a similar number of solutions. The hypervolume convergence curves in Fig.~\ref{fig:surrogate_model}\subref{fig7:sub2} reveal that the surrogate model efficiently steers the search toward promising regions of the vast solution space. This demonstrates that surrogate assistance is crucial for efficiently discovering superior merging configurations.

\subsubsection{Search Strategy}

\begin{figure}[htbp]
\centering
\subfloat[]{%
    \includegraphics[width=0.48\linewidth]{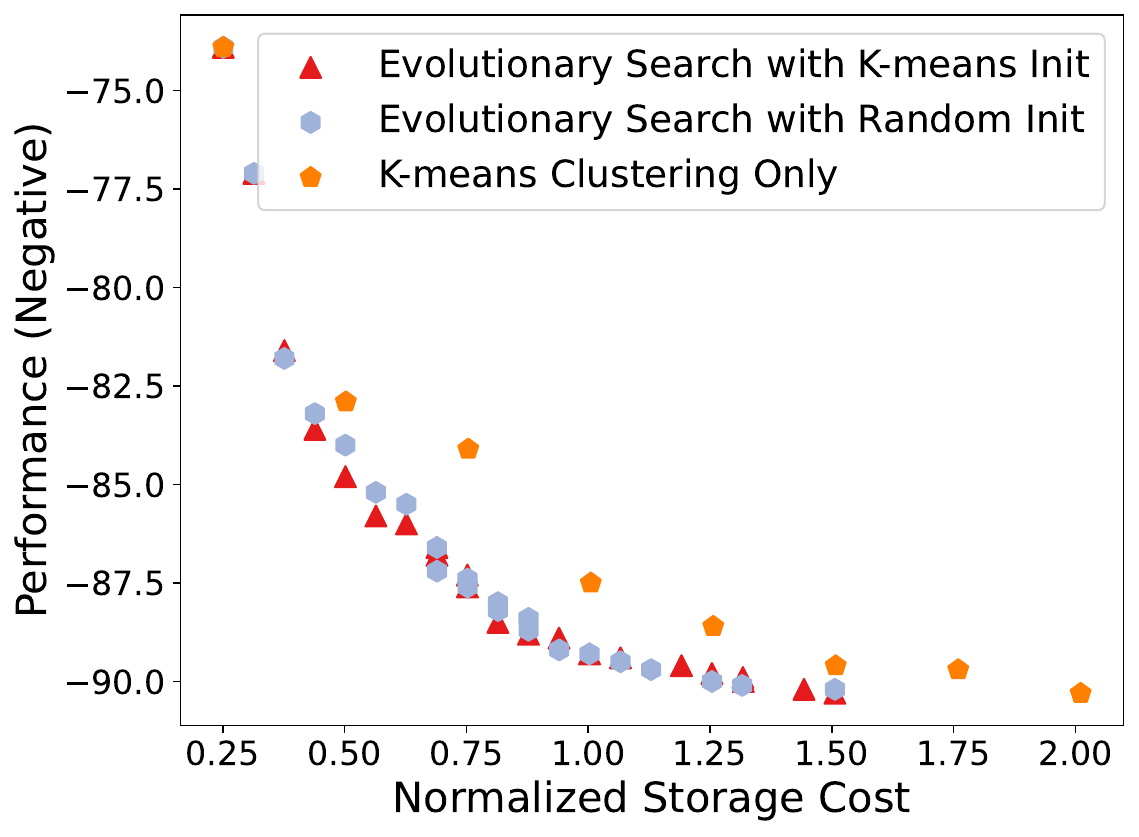}%
    \label{fig8:sub1}}
\hfil
\subfloat[]{%
    \includegraphics[width=0.48\linewidth]{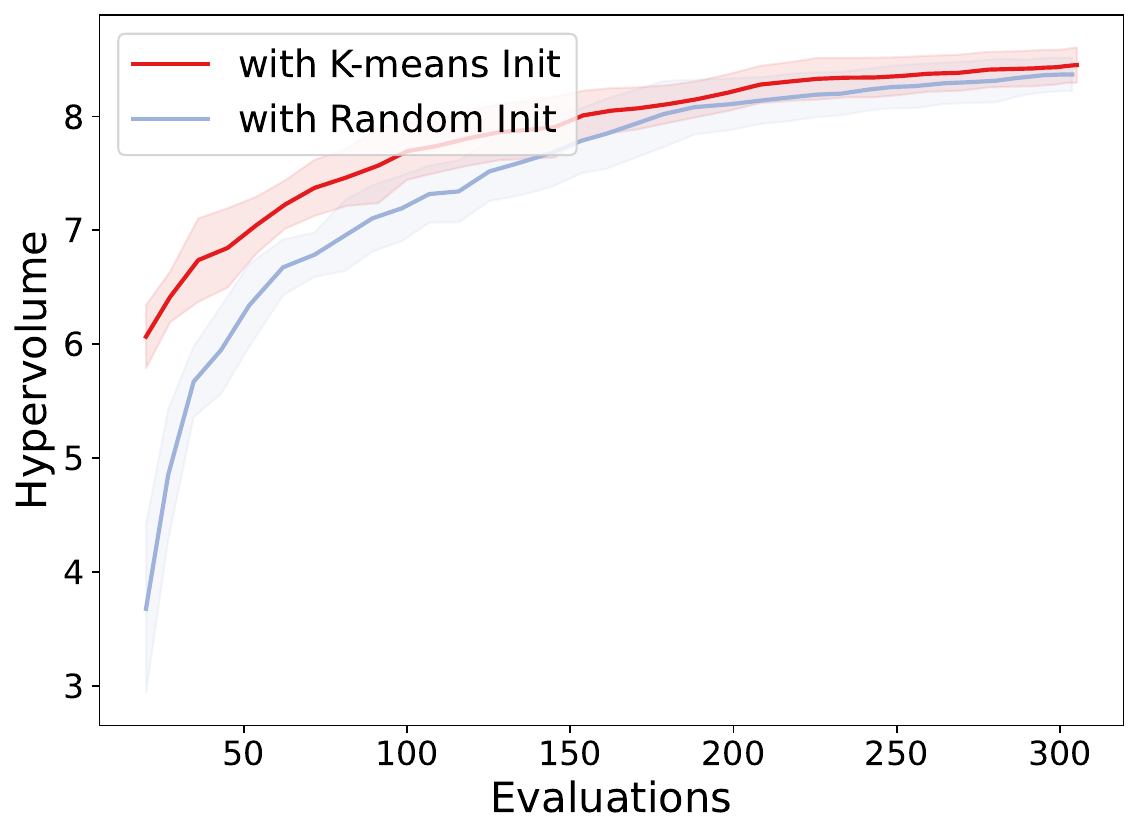}%
    \label{fig8:sub2}}
\caption{The impact of search strategy in MERGE. (a) Pareto-optimal solutions. (b) Hypervolume evolution.}
\label{fig:grouping_strategy}
\end{figure}

Fig.~\ref{fig:grouping_strategy}\subref{fig8:sub1} demonstrates that our evolutionary search with K-means initialization yields a superior Pareto front compared to random initialization. While the K-means-only approach is inadequate, as its solutions are entirely dominated, it serves as an effective enhancement to the evolutionary search. The HV convergence trend (Fig.~\ref{fig:grouping_strategy}\subref{fig8:sub2}) further validates that incorporating the K-means prior accelerates the discovery of high-quality solutions, thereby improving optimization efficiency.

\section{Conclusion}
\label{sec:Conclusion}

In this paper, we address the fundamental limitations of existing model merging techniques, specifically the inefficiency of instance-specific merging and the neglect of component-wise heterogeneity, by proposing MERGE. This novel approach systematically reconciles the tension between adaptability and reusability through modular expert recombination, achieving both component-wise and input-aware fine-grained merging. 

By decoupling the merging process into an offline optimization stage and an on-demand reuse stage, MERGE constructs a bridge between high-quality configuration search and efficient inference deployment. During offline optimization, we formulate component-wise merging as a bi-objective optimization problem to balance cross-task performance and storage efficiency, and develop a surrogate-assisted evolutionary algorithm to efficiently explore the vast, discrete search space. This process yields a diverse set of Pareto-optimal merging configurations that underpin a reusable modular expert library. During on-demand reuse at inference time, MERGE leverages a lightweight routing network to dynamically retrieve and recombine these offline-derived modular experts from the library, enabling the assembly of input-specific models under user-specified resource constraints. 

MERGE enables more adaptive and flexible model merging, consistently outperforming baselines across diverse experimental settings and domains, including varying model scales (e.g., ViT-B/32, ViT-L/14), task types (e.g., vision, language), and fine-tuning strategies (e.g., FFT, PEFT). Our systematic analysis further uncovers critical insights into the intrinsic mechanisms of fine-grained model merging.

% Looking ahead, several potential directions merit further exploration. First, more advanced optimization and surrogate-assisted techniques could be developed to further enhance the efficiency and precision of the multi-objective component-wise merging. Second, future endeavors could extend MERGE beyond fine-grained merging of homogeneous models to systematically handle heterogeneous models, thereby further broadening its applicability. Finally, to establish a stronger theoretical foundation for fine-grained knowledge integration, future research could investigate the intrinsic mechanisms of fine-grained model merging, thereby improving its interpretability and guiding the design of more principled merging strategies.

% \section*{Acknowledgments}

\bibliographystyle{IEEEtran}
\bibliography{IEEEabrv,TNNLS_MERGE}
\end{document}